\documentclass[sigconf]{acmart}

\usepackage{flushend}
\usepackage{subfigure,multirow}
\usepackage{graphicx}
\usepackage{xurl}
\usepackage{color}

\usepackage{array}
\usepackage{amsmath}
\usepackage{mathrsfs} 
\usepackage{autobreak}
\usepackage{algorithm}
\usepackage[noend]{algorithmic}
\usepackage{dsfont}

\begin{document}
	

	\acmYear{2023}\copyrightyear{2023}
	\setcopyright{acmlicensed}
	\acmConference[MobiSys '23]{ACM International Conference on Mobile Systems, Applications, and Services}{June 18--22, 2023}{Helsinki, Finland}
	\acmBooktitle{ACM International Conference on Mobile Systems, Applications, and Services (MobiSys '23), June 18--22, 2023, Helsinki, Finland}
	\acmPrice{15.00}
	\acmDOI{10.1145/3581791.3596852}
	\acmISBN{979-8-4007-0110-8/23/06}

	\ccsdesc[500]{Computer systems organization~Embedded and cyber-physical systems}
	\ccsdesc[500]{Computing methodologies~Artificial intelligence}
	
	\keywords{On-Device Training, Neural Network, Speedup, Tensor Selection, Elasticity}
	
	\title[]{ElasticTrainer: Speeding Up On-Device Training with Runtime Elastic Tensor Selection}
	
	\author{Kai Huang}
	\affiliation{%
		\institution{University of Pittsburgh}
		\country{USA}
	}
	\email{k.huang@pitt.edu}
	
	\author{Boyuan Yang}
	\affiliation{%
		\institution{University of Pittsburgh}
		\country{USA}
	}
	\email{by.yang@pitt.edu}
	
	\author{Wei Gao}
	\affiliation{%
		\institution{University of Pittsburgh}
		\country{USA}
	}
	\email{weigao@pitt.edu}
	
	\begin{abstract}	
		On-device training is essential for neural networks (NNs) to continuously adapt to new online data, but can be time-consuming due to the device's limited computing power. To speed up on-device training, existing schemes select trainable NN portion offline or conduct unrecoverable selection at runtime, but the evolution of trainable NN portion is constrained and cannot adapt to the current need for training. Instead, runtime adaptation of on-device training should be fully elastic, i.e., every NN substructure can be freely removed from or added to the trainable NN portion at any time in training. In this paper, we present \emph{ElasticTrainer}, a new technique that enforces such elasticity to achieve the required training speedup with the minimum NN accuracy loss. Experiment results show that ElasticTrainer achieves up to 3.5$\times$ more training speedup in wall-clock time and reduces energy consumption by 2$\times$-3$\times$ more compared to the existing schemes, without noticeable accuracy loss.		
	\end{abstract}
	
	\maketitle
	
	\section{Introduction}
	Neural Networks (NNs) have been widely used on mobile and embedded devices for image and speech recognition \cite{amodei2016deep, hu2015face}. In these applications, NNs are pre-trained offline with large datasets (e.g., ImageNet \cite{deng2009imagenet} and BooksCorpus \cite{zhu2015aligning}) before on-device deployment, but pre-trained NNs may not be able to capture the new patterns of online data. For example, a NN pre-trained with the ImageNet dataset for pedestrian detection suffers 20\% inference accuracy drop on the PASCAL dataset \cite{torralba2011unbiased,everingham2010pascal}\footnote{The PASCAL dataset is another widely used dataset for object detection and recognition \cite{everingham2010pascal}.}. On-device training with online data, hence, is essential for NNs to retain generality or be personalized \cite{zhang2020mdldroidlite, gim2022memory, lin2022device}. 
 
	
	In particular, in many applications such as drone search \& rescue \cite{turki2022mega}, personalized facial identification \cite{li2020look} and embodied AI-driven robotics \cite{kumar2021rma, smith2022legged}, it is imperative that on-device NN models are updated within short time, so as to promptly adapt to the new online data patterns or real-time feedback from the environment. However, on-device training of NN models is usually time-consuming due to the devices' limited computing power. For example, using a Raspberry Pi 4 \cite{rpi4} to train a ResNet50 \cite{he2016deep} model with the CUB-200 dataset \cite{WahCUB_200_2011} that contains 5,994 images can take $>$30 days. Even on stronger devices with GPUs such as Nvidia Jetson TX2 \cite{jetsontx2}, the training on each image could still take 30 seconds.

	\begin{figure}
		\centering
		\includegraphics[width=0.65\linewidth]{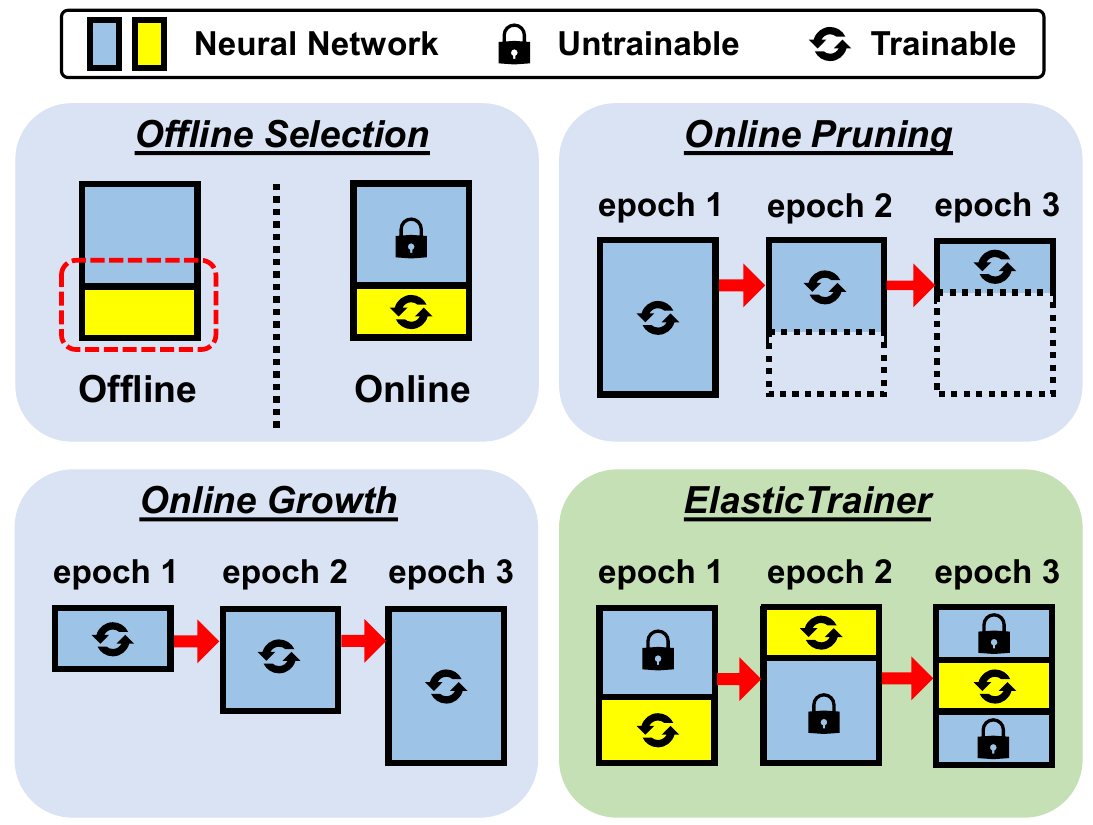}
		\vspace{-0.1in}
		\caption{Existing work vs. ElasticTrainer}
		\vspace{-0.2in}
		\label{fig:comparison}
	\end{figure}
	
	To reduce such long delay of on-device training, one could opt to use a smaller NN model, but may suffer from the impaired model performance when being applied to new online data, due to its limited learning power. Another alternative is to offload NN training to the cloud \cite{vepakomma2018split, poirot2019split}, but incurs high communication overhead. For example, training a VGG16 \cite{simonyan2014very} model with the CUB-200 dataset requires transmitting $>$10GB data for each epoch. Even with high-throughput network connections, it would be still difficult for cloud servers to timely handle large amounts of retraining requests from the large population of mobile and embedded devices.



	Instead, to speed up the training at the local device, an intuitive solution is to leverage hardware accelerators such as specialized DSP \cite{xu2022mandheling}) or NPU \cite{jeong2022band}, but their availability is limited on mobile and embedded devices. Traditional transfer learning \cite{donahue2014decaf, sharif2014cnn} (Figure \ref{fig:comparison} top-left) suggests  only training a portion of the NN model. It hypothesizes that most top layers have learned the generic capabilities of feature extraction, and restricts the trainable portion to the few bottom layers \cite{chatfield2014return}. This restriction, however, weakens NN's learning power and reduces its inference accuracy by 20\% on difficult learning tasks \cite{cai2020tinytl}. Later studies remove this restriction by identifying important NN substructures (e.g., bias parameters \cite{zaken2021bitfit, lin2022device}, normalization layers \cite{mudrakarta2018k} and small parallel branches \cite{hu2021lora}), but only evaluate such importance offline without considering the variability of online data that result in dynamic training feedback \cite{akata2015evaluation}. Hence, online importances of NN substructures could be largely different from those offline, leading to noticeable accuracy loss.

	
	

	A better choice is to adaptively adjust the trainable NN portion at runtime. NN pruning \cite{he2017channel, molchanov2019importance} for on-device training removes less important NN structures on the fly \cite{lym2019prunetrain, jiang2022model} (Figure \ref{fig:comparison} top-right). However, since the pruned NN portions can never be selected again even if they may be useful \cite{kemker2018measuring}, NN's representation power is weakened over time and becomes insufficient for difficult learning tasks. Training can also start from a small NN that gradually grows \cite{irsoy2019continuously, zhang2020mdldroidlite} (Figure \ref{fig:comparison} bottom-left), but the newly added NN layers are trained from scratch and require $>$30\% extra training epochs \cite{zhang2020mdldroidlite}. 

	The key reason for the existing runtime adaptation approaches to fail is that their selections of the trainable NN portion are one-way and unrecoverable processes. The evolution of trainable NN portion is hence constrained and cannot flexibly adapt to the current need for training. Instead, we envision that such runtime adaptation should be fully \emph{elastic}, i.e., every NN substructure can be freely removed from or added to the trainable NN portion at any time in training, as needed. In this paper, we present \emph{ElasticTrainer} (Figure \ref{fig:comparison} bottom-right) to enforce such elasticity at the granularity of NN tensors, and aim to achieve the required training speedup with the minimum NN accuracy loss. ElasticTrainer evaluates the importance of NN tensors in different training stages at runtime, and adaptively selects the smallest set of important tensors to achieve the required training speedup. 
	
	
	ElasticTrainer's selection of trainable NN portion builds on appropriate evaluation of NN tensors' importance, which is challenging because tensors do not directly associate with any input data variables or intermediate features. As a result, traditional approaches based on weight magnitudes \cite{li2016pruning}, random perturbations \cite{breiman2001random} or attention \cite{bahdanau2014neural, vaswani2017attention,guo2020adafilter} will be either inaccurate or expensive, and most eXplainable AI (XAI) techniques that are based on attribution \cite{selvaraju2017grad,sundararajan2017axiomatic} are not applicable. Instead, our approach is to follow the similar rationale with current XAI techniques that measure the importance of an input data variable as the accumulation of relevant gradients, to evaluate tensor importance as the cumulative gradient changes of its weight updates in training. In this way, we ensure that selected tensors will make the maximum contribution to reducing the training loss.
	
	
	Based on such importance evaluation, we aim to achieve training speedup in wall-clock time instead of FLOPs (number of floating point operations) that is widely used in the existing work \cite{lym2019prunetrain, zhang2020mdldroidlite}. The major reason is that using FLOPs ignores the hardware accelerations for different NN operations. For example, convolution layers take $>$95\% FLOPs in ResNet50 model's training, but only take $<$60\% of wall-clock time when running on Nvidia GPUs, due to the TensorCore's acceleration for convolution. Ignoring these practical factors could result in large difference between training speedups in theory and practical settings.
	
	The main challenge of achieving wall-clock time speedup is how to precisely profile the training times of different tensor selections. Due to the interdependency between training times of selected tensors, the total training time of selected tensors is not equal to the summation of tensors' individual training times. To tackle this challenge, we build a new time model that incorporates the relations between tensors and NN operations into the training time profiling. Based on this model, we develop a dynamic programming (DP) algorithm that can find the optimal tensor selection from the exponential number of possibilities (e.g., $2^{214}$ for 214 tensors in ResNet50 model \cite{he2016deep}), with negligible computing overhead.
	
	
	
	To our best knowledge, ElasticTrainer is the first work that allows full elasticity in on-device training, which is essential to achieve training speedup for difficult learning tasks (e.g., fine-grained image recognition \cite{WahCUB_200_2011,khosla2011novel}). Our detailed contributions are as follows:
	\begin{itemize}
		\item We leverage the rationale of XAI for runtime evaluation of tensor importance in training with high accuracy and adaptability.
		\item We build new time models that allow precise profiling of tensor selection's training execution time.
		\item Our lightweight DP algorithm can decide the optimal selection of tensors at runtime that maximizes the training loss reduction.
	\end{itemize}
	
	We implemented ElasticTrainer on multiple embedded devices including Nvidia Jetson TX2 and Raspberry Pi 4B\footnote{Our source codes can be found at \url{https://github.com/HelloKevin07/ElasticTrainer}.}, and evaluated its performance on various popular datasets. From our experiment results, we have the following conclusions:
	\begin{itemize}
		\item ElasticTrainer is \emph{time efficient}. Compared to the existing schemes \cite{donahue2014decaf, sharif2014cnn, cai2020tinytl, mudrakarta2018k, lym2019prunetrain}, it achieves up to 3.5$\times$ more training speedup in wall-clock time and reduces the training FLOPs by 60\%. 
		\item ElasticTrainer is \emph{accurate}. It incurs negligible accuracy loss of NN prediction compared to full training on most datasets we used, and achieves at least 10\% accuracy improvement compared to the existing schemes, especially on difficult datasets such as CUB-200.	
		\item ElasticTrainer is \emph{adaptive}. It is able to achieve different speedup objectives and maximize the NN accuracy in different datasets, training stages and NN models.
		\item ElasticTrainer is \emph{lightweight}. Its selection of trainable NN portion incurs $<$1\% extra computation, and reduces the energy consumption of training by 2$\times$-3$\times$ more compared to the existing schemes.
	\end{itemize}
	
	\vspace{-0.05in}
	\section{Background \& Motivation}
	To help better understand our design of ElasticTrainer, we first demonstrate the opportunities of speeding up on-device NN training with small accuracy loss, hence motivating our elastic selection of the trainable NN portion at runtime. Afterwards, to optimize such speedup, we describe the time model of NN training and discuss our choices of granularity in selecting the trainable NN portion.
		
	\vspace{-0.05in}
	\subsection{Opportunities for Training Speedup}
	Since the pre-trained NN model has learned generic capabilities of extracting low-level features (e.g., color and texture information in images \cite{bengio2012deep}), on-device training only needs to be applied to some NN substructures and hence requires fewer training epochs and weight updates \cite{chen2015net2net} compared to training the model from scratch. As a result, we can potentially gain significant speedup by selecting a small trainable NN portion without losing much accuracy. 

\begin{figure}[ht]
		\centering
		\vspace{-0.15in}
		\hspace{-0.25in}
		\subfigure[Training progress and latency in the first 3 epochs] { 
			\includegraphics[width=0.24\textwidth]{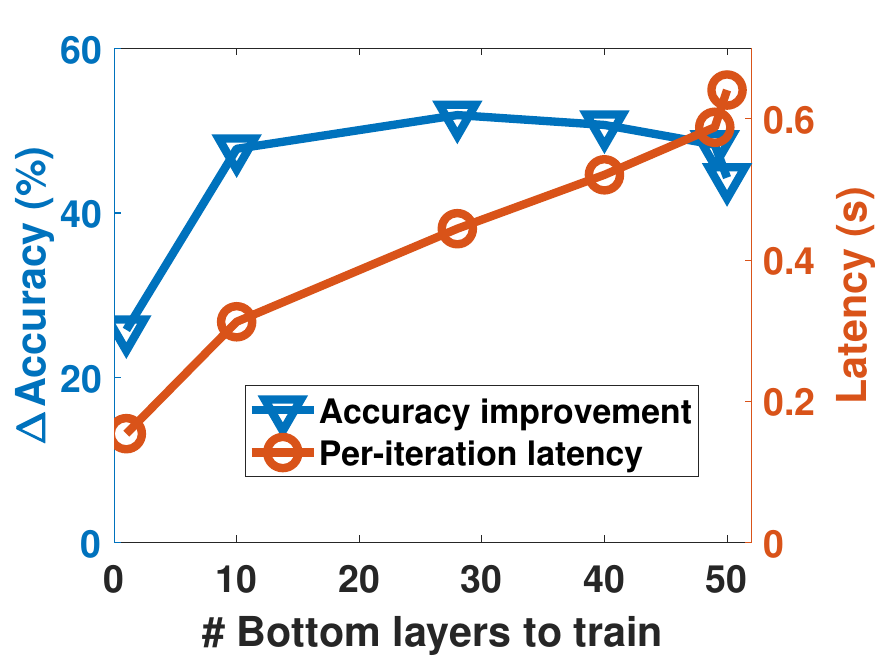}
			\label{fig:opportunity_1}
		}
		\subfigure[Suboptimality of existing work in selecting the trainable NN portion] { 
			\includegraphics[width=0.24\textwidth]{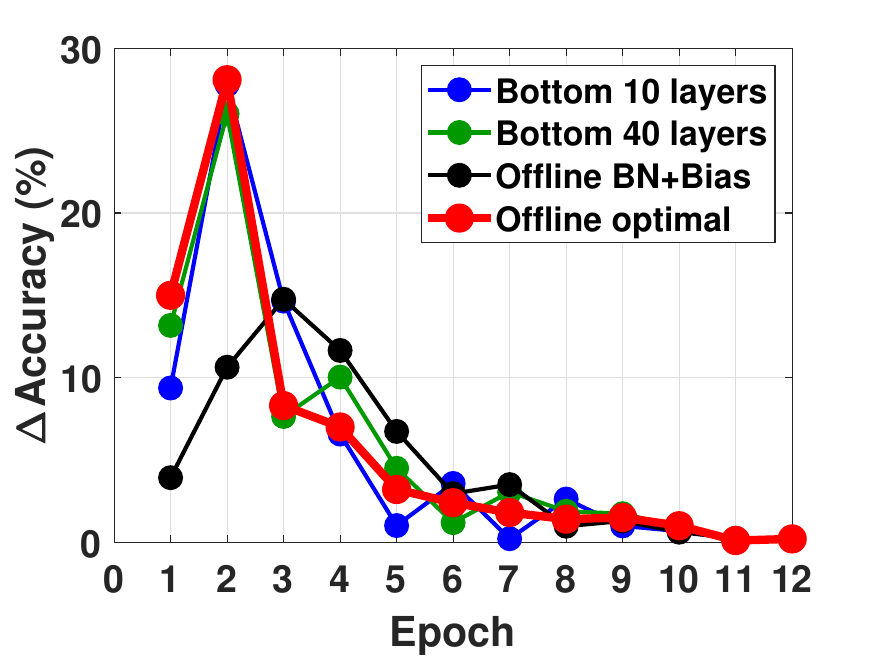}
			\label{fig:opportunity_2}
		}
		\hspace{-0.4in}
		\vspace{-0.15in}
		\caption{Opportunities of training speedup}
		\label{fig:opportunity}
		\vspace{-0.1in}
	\end{figure}

	
	To verify this, we use a ResNet50 NN model being pre-trained with the ImageNet dataset and retrain it with the CUB-200 dataset. The retraining adopts the traditional transfer learning methods \cite{chatfield2014return} and selects the trainable NN portion as a number of bottom NN layers. Experiment results on an Nvidia Jetson TX2 in Figure \ref{fig:opportunity_1} show that, when the trainable NN portion only contains 10 bottom layers, the training progress in the first 3 epochs, measured as the improvement of validation accuracy, is similar to that of training the entire NN (50 layers), but with a 2x training speedup.

	These traditional learning methods, however, cannot ensure optimal selection of the trainable NN portion. To explore their suboptimality, we set an objective of 1.7x training speedup and exhaustively search offline for the optimal trainable NN portion in the pre-trained ResNet50 model. Results in Figure \ref{fig:opportunity_2} show that, when retraining with the CUB-200 dataset starts in epoch 1, compared to this optimal portion, no matter if the trainable NN portion is selected from the bottom NN layers, NN weights in batch normalization layers or bias weights in convolutional and dense layers (Offline BN+Bias), it cannot adapt well to the new dataset.
	
	On the other hand, such optimal selection of trainable NN portion may also vary in different training stages at run-time. As shown in Figure \ref{fig:opportunity_2}, although the offline optimal selection ensures the best training progress in the first 2 epochs, its optimality quickly deteriorates as training further proceeds\footnote{Although traditional learning methods outperform the offline optimal selection after epoch 3, we have no clue if any of these selections is optimal.}, verifying that importances of NN structures could vary in different training epochs. These results, hence, demonstrate the ineffectiveness of pruning-based methods in on-device training, and instead motivate us to transform the online selection of trainable NN portion from one-way and unrecoverable to fully elastic.

	\begin{figure}[ht]
		\centering
				\vspace{-0.1in}
		\includegraphics[width=0.95\linewidth]{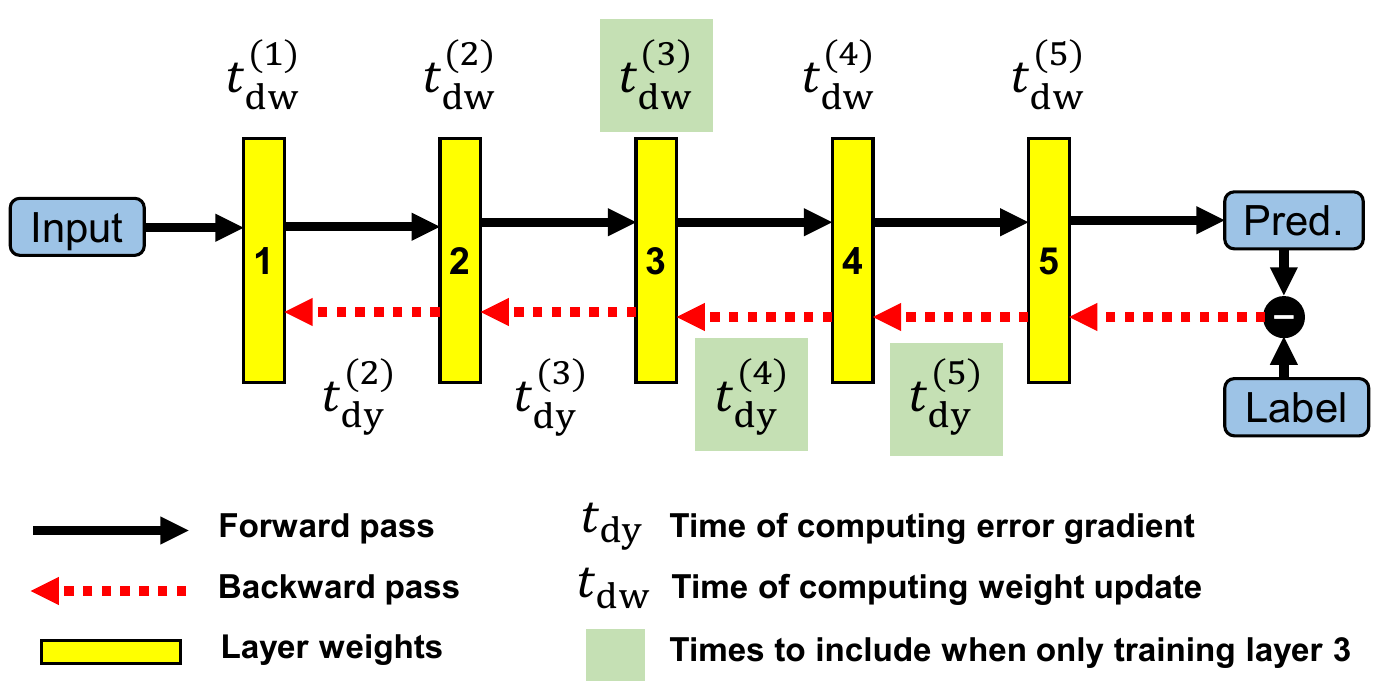}
		\vspace{-0.1in}
		\caption{Forward \& backward passes of NN training}
		\vspace{-0.1in}
		\label{fig:backpropagation}
	\end{figure}
	
	\vspace{-0.05in}
	\subsection{Time Model of NN Training}
	\label{subsec:time_model}
	Our objective of ElasticTrainer is to achieve the desired training speedup in wall-clock time, and we hence need to incorporate the corresponding time model of NN training into the selection of trainable NN portion. As shown in Figure \ref{fig:backpropagation}, the training of most existing NNs consists of forward and backward passes \cite{hecht1992theory}. In a forward pass, the NN takes a batch of training data as input and extracts features on a layer basis to produce final predictions. A loss function is then computed by comparing the predictions with labels. Its training time, hence, is the cumulation of all layers' computing times of such feature extraction. In a backward pass, the NN weights are recursively updated on a layer basis based on the error gradient feedback \cite{amari1993backpropagation} computed from the loss value. As a result, in the backward pass, layer $i$ spends time $t_{dw}^{(i)}$ to compute the weight update using the error gradient passed from layer $i+1$, and spends time $t_{dy}^{(i)}$ to compute the error gradient being passed to layer $i-1$. 
	
	Even if a layer is not selected, it still needs to compute and pass error gradients, and the time needed for training hence do not only depend on the selected layers. For example in Figure \ref{fig:backpropagation}, even if only layer 3 is selected, the times for computing error gradients in layer 4 and 5 should still be counted towards the total time for training, which is then calculated as $t_{dw}^{(3)} + t_{dy}^{(4)} + t_{dy}^{(5)}$. Since the gradient computations in backward passes are much more time consuming than forward passes, incorporating these times of gradient computations is important to ensure accurate estimation of NN training time and correct selection of trainable NN portion. In the rest of this paper, our tensor selection will also be mainly focusing on reducing the training time in backward passes.

	\begin{figure}[ht]
		\centering
		\vspace{-0.1in}
		\includegraphics[width=0.86\linewidth]{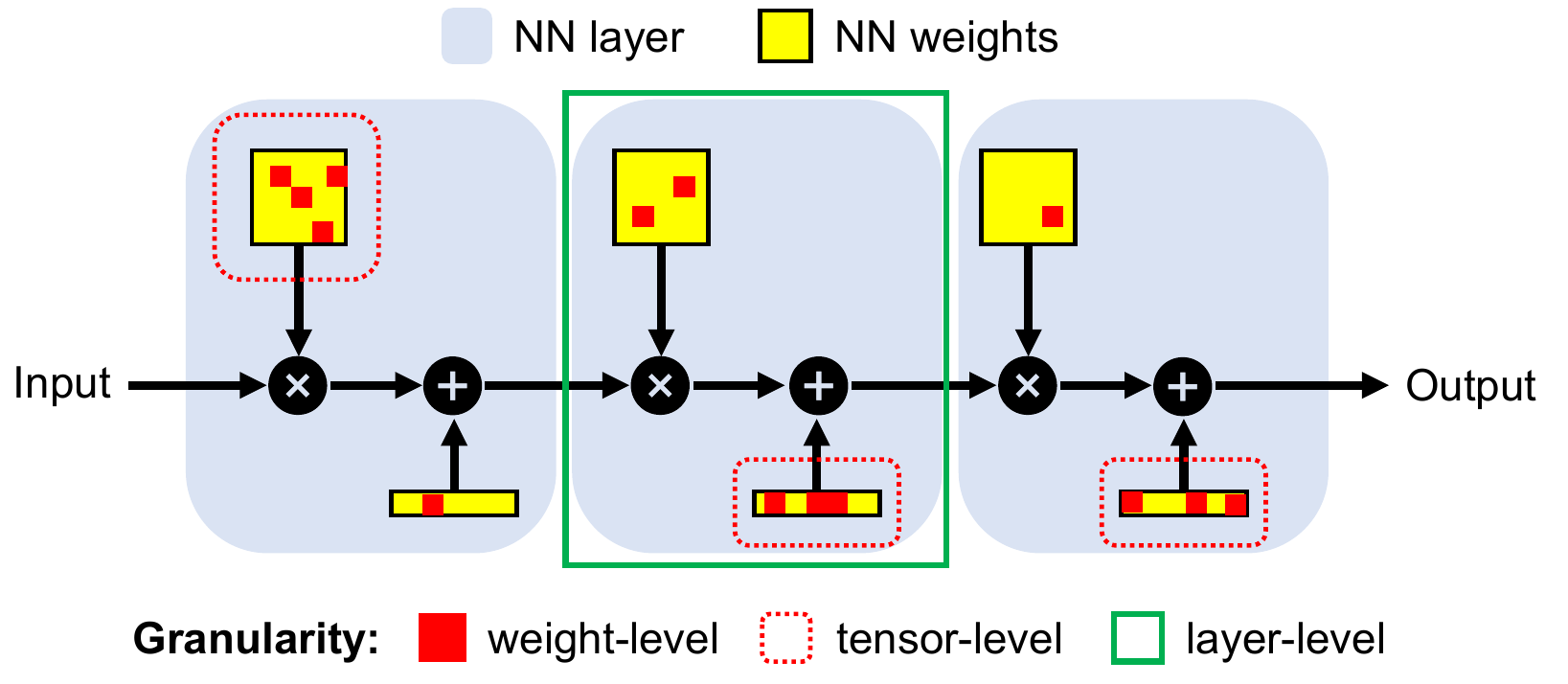}
		\vspace{-0.1in}
		\caption{Granularities of selection}
		\vspace{-0.15in}
		\label{fig:selection_granularity}
	\end{figure}

	\begin{figure*}[ht]
		\centering
		\includegraphics[width=5.75in]{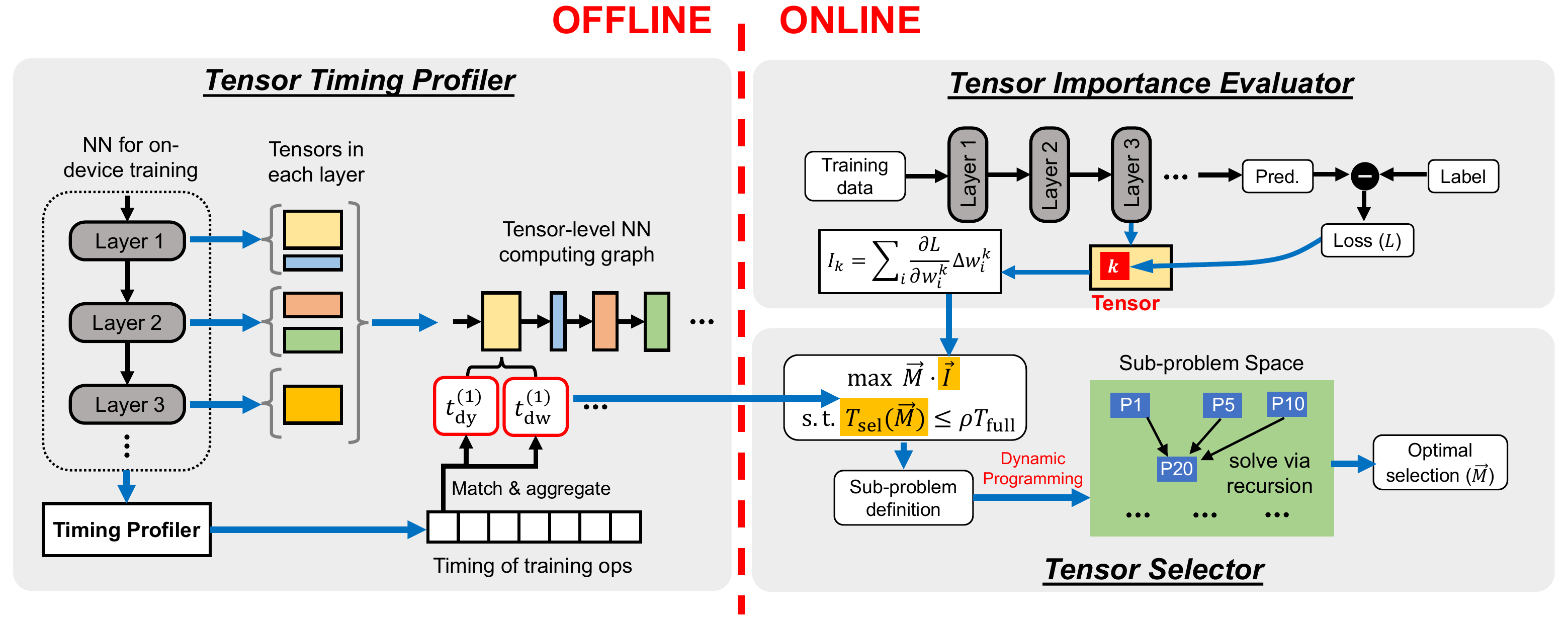}
		\vspace{-0.15in}
		\caption{Overview of ElasticTrainer Design}
				\vspace{-0.15in}
		\label{fig:system_overview}
	\end{figure*}
	
	\vspace{-0.05in}
	\subsection{Granularity of Selection}
	The optimality of selecting the trainable NN portion, however, is affected by the granularity of selection, which can be at the level of weights, tensors and layers on most NNs (e.g., convolutional and dense NNs). As shown in Figure \ref{fig:selection_granularity} with a three-layer dense NN as an example, each layer contains two trainable tensors\footnote{In this paper, we use tensor as a generic term for vector, matrix, and other higher dimensional data forms.} as groups of weights, and they apply multiplication and addition to the layer input, respectively. 
	
	Among these granularities for selection, the layer-level selection is coarse-grained and hence inaccurate because some important weights within a layer may not be selected due to many others that are unimportant. On the other hand, although the weight-level selection is the most fine-grained, it requires fine-grained indexing with irregular data access patterns, which could lead to poor training efficiency on existing NN software frameworks such as TensorFlow \cite{abadi2016tensorflow} and PyTorch \cite{paszke2019pytorch} with vector and matrix based implementations. Hence, ElasticTrainer adopts tensor-level selection, which ensures accuracy and can also be efficiently executed in existing NN frameworks without extra overhead. 
	
	In particular, to adopt the layer-level time model of NN training described in Section \ref{subsec:time_model}, tensors can be equivalently considered as smaller layers that are sequentially connected based on their characteristics. By incorporating this tensor-level time model, ElasticTrainer first constructs a selection-time model which reflects the relationship between the selected tensors and NN training time. It then uses such a model to formulate an optimization problem to find the optimal selection of trainable NN portion.

	\vspace{-0.05in}
	\section{Overview}
	
	To speed up on-device training, an intuitive formulation of the speedup problem is to minimize the training time with respect to the required NN training quality, which is measured by the trained NN model's prediction accuracy. However, it is practically hard to decide in advance the appropriate requirement of such accuracy that can be achieved within reasonable training time. Instead, as shown in Figure \ref{fig:system_overview}, ElasticTrainer's design aims to select the optimal trainable NN portion at runtime, to achieve the desired training speedup with the maximum training loss reduction\footnote{In our problem formulation, we use the amount of training loss reduction to evaluate the improvement of training quality, which indirectly measures the NN model's prediction accuracy loss in inference.}. 
	
	This selection problem is formulated as a constrained optimization problem as follows:
	\begin{equation}
	\max \ \Delta_{loss}(\mathcal{M}) \ \ \ \ \text{s.t.} \ \ T_{selective}(\mathcal{M}) \le \rho T_{full},
	\label{eq:formulation}
	\end{equation}
	where $\mathcal{M}$ is a binary mask to be solved denoting tensor selection. $T_{selective}$ is the estimated training time according to the time model in Section \ref{subsec:time_model}, and it is constrained to be lower than a user-specified ratio ($\rho$) of the full training time ($T_{full}$), as the objective of training speedup. For example, $\rho$=50\% means that the training time should be reduced to 50\% of that in full training. The time needed for full training, on the other hand, can be estimated in advance from the training parameters, such as the computing time per iteration, batch size, dataset size and number of training epochs. In practice, users can either set $\rho$ to be constant or adjust it at runtime, and this decision only relates to the application's requirement on timeliness.

	In the offline stage, ElasticTrainer uses a \emph{Tensor Timing Profiler} to profile the training times of selected tensors, to provide inputs for calculating $T_{selective}(\mathcal{M})$. In the online stage, solving the problem in Eq. (\ref{eq:formulation}) builds on an accurate yet computationally efficient metric that evaluates the aggregate importance of  selected tensors and the corresponding reduction of training loss, and such evaluation is done by \emph{Tensor Importance Evaluator} in ElasticTrainer design. The outputs of these two modules are used by \emph{Tensor Selector} to solve the selection problem via dynamic programming.
	
	\vspace{-0.05in}
	\subsection{Tensor Importance Evaluator}
	\label{subsec:overview_tensor_importance}
	An intuitive approach to evaluating the importance of NN structures is based on their magnitudes \cite{li2016pruning}, but cannot accurately reflect the interdependency of different weight updates in backward passes. Hence, they cannot ensure correct selection of tensors that contain multiple interdependent weights. Instead, eXplainable AI (XAI) techniques suggest using gradient-based methods to incorporate such dependencies \cite{selvaraju2017grad, sundararajan2017axiomatic}. However, most of these methods are limited to input data variables or intermediate features. 
	
	
	To address these limitations, we leverage the similar rationale of existing XAI approaches, and extend these approaches to evaluate the importance of tensors from the gradient changes of their weight updates. We define the importance of a NN tensor $k$ in a specific training epoch as:
	\vspace{-0.05in}
	\begin{equation}
	I_k = \sum_{i} \frac{\partial L}{\partial w_i^k} \Delta w_i^k,
	\vspace{-0.05in}
	\label{eq:importance}
	\end{equation}
	where $L$ and $w_i^k$ denote the training loss function and the $i$-th weight in tensor $k$, respectively, and $\Delta w_i^k$ is the recent update of $w_i^k$ in the training epoch. This metric aggregates how each weight update contributes to the reduction of training loss, and its gradient computation mimics the similar computation in backward pass to naturally incorporate the impact of weight dependencies in training. 
	
	The prerequisite of correctly using this importance metric to select the trainable NN portion is that this metric should satisfy additivity, which ensures that the importances of multiple weights in a tensor can be correctly aggregated. To achieve such additivity, in the practical training process, we take first-order approximation to the importance evaluation specified in Eq. (\ref{eq:importance}), and the detailed rationale of such approximation is described in Section \ref{sec:importance_evaluation}.
	
	To ensure that the importance evaluation always aligns with the current training progress, we will need to frequently evaluate tensor importance at run-time. In ElasticTrainer, we set such period of importance evaluation to be a few training epochs by assuming that the tensor importance remains constant in each short period, and then use the weight update in the first epoch of each period for importance evaluation. The impact of such period length on the training efficiency and accuracy will also be evaluated in Section \ref{sec:importance_evaluation}.
	

	\begin{figure}[ht]
		\centering
		\vspace{-0.1in}
		\includegraphics[width=0.86\linewidth]{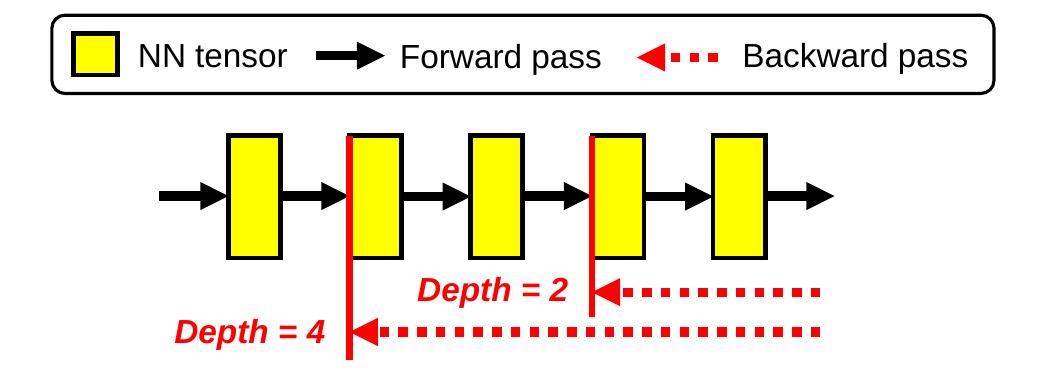}
		\vspace{-0.1in}
		\caption{Problem decomposition with constrained depth of backward pass}
		\vspace{-0.1in}
		\label{fig:backward_depth}
		\vspace{-0.05in}
	\end{figure}
	
	\vspace{-0.05in}
	\subsection{Tensor Timing Profiler}
	Standard NN profilers\footnote{For example, TensorFlow provides a built-in profiler: https://www.tensorflow.org/guide/profiler} can measure the execution time of each NN operation (e.g., matrix multiplication and convolution) in offline training. However, such measurements are limited to the operation level and have no clear correspondence to NN tensors that participate in these operations. In other words, when a set of selected tensors is being trained, the execution time of the involved NN operations will not equal to the summation of training times of individual tensors. To address this limitation, we first convert the original layer-based NN structure into a tensor-level computing graph, which retains the execution order of all tensors' involvements in training. Then, we compute the backward pass timing $t_{dy}$ and $t_{dw}$ of each tensor by aggregating the timings of related NN operations from the standard profiler.
	
	The key challenge is how to determine the execution order of tensors and how to aggregate the timings of related NN operations, both of which depend on the type of NN layers where tensors are located. We take different types of NN layers (e.g., convolutional, dense, and batch normalization layers) into consideration and develop rules for profiling each type of layer. Details of such profiling are in Section \ref{sec:profiling}.

	\vspace{-0.05in}
	\subsection{Tensor Selector}
	Intuitively, selecting the top-$k$ of most important tensors will change the training speed at every interval. However, it is hard to estimate the cumulative training time or ensure that the required speedup can be met. In contrast, we use the timing profiles $(\vec{t_{dy}}, \vec{t_{dw}})$ and importance $\vec{I}$ of tensors to instantiate the objective and constraint in Eq. (\ref{eq:formulation}). With a binary mask $\vec{M}$ where its $j$-th element indicates if tensor $j$ is selected, Eq. (\ref{eq:formulation}) can be rewritten as 
	\begin{equation}
	\max \ \vec{M} \cdot \vec{I} \ \ \ \ \ \text{s.t.} \ \ T_{forward} + \vec{M} \cdot \vec{t_{dw}} + f(\vec{M}) \cdot \vec{t_{dy}} \le \rho T_{full},
	\label{eq:detailed_formulation}
	\end{equation}
	where $T_{forward}$ indicates the training time of forward passes that is fixed for any tensor selection, and $f(\vec{M})$ is another binary mask indicating the tensors whose times of computing error gradient along backward pass ($t_{dy}$) should be included in the training time. According to the time model in Section 2.2, only $t_{dw}$ of selected tensors are included in training time, but for any selected tensor, all $t_{dy}$ of later layers should be included. For example, if $\vec{M}=[0,0,0,1,1,0,0]$, then $f(\vec{M})=[0,0,0,1,1,1,1]$. Tensor Solver aims to solve the selection mask $\vec{M}$ that maximizes the importance of selected tensors while restraining the training time within $\rho T_{full}$.
	
	Since Eq. (\ref{eq:detailed_formulation}) is a nonlinear integer programming problem and hence NP-hard \cite{wang2011approximation}, we will solve it in pseudo-polynomial time by dynamic programming (DP). Specifically, as shown in Figure \ref{fig:backward_depth}, we decompose the whole problem into many subproblems which are constrained by different depths of backward pass. These subproblems can be sequentially solved from the easiest one with the smallest depth, by using their recurrence relations. Details of our DP algorithm design and analysis are in Sec 6.
	
	
	\vspace{-0.05in}
	\section{Evaluating Tensor Importance}
	\label{sec:importance_evaluation}
	Since NN training iteratively updates the NN weights to minimize the loss function, an intuitive approach to evaluating the importance of a weight update $\Delta w$ in a given training epoch is to undo this update and check how the training loss value increases back:
	\vspace{-0.05in}
	\begin{equation}
	\Delta L = L(w) - L(w + \Delta w),
	\vspace{-0.05in}
	\label{eq:naive_importance_computation}
	\end{equation}
	so that higher $\Delta L$ means this update is more important. However, repeatedly applying this approach to every NN weight is expensive due to the large number of NN weights. For example, a ResNet50 model with 50 NN layers contains $>$23 millions of weights. 
	
	Instead, our approach is to leverage the gradient computation that has been widely used in backward pass in training and can be efficiently done with auto-differentiation software \cite{baydin2018automatic}. More specifically, we approximate the importance evaluation in Eq. (\ref{eq:naive_importance_computation}) by smoothing the undo operation and computing the loss gradients with respect to the updates corresponding to all the NN weights. Letting the multiplicative $\vec{c} \in [0, 1]^{M}$ denote the continuous undo operation for all the $M$ NN weights, we can compute the loss gradient with respect to $\vec{c}$ as
	\vspace{-0.05in}
	\begin{equation}
	\frac{\partial L(\vec{w} + \vec{c} \odot \Delta \vec{w})}{\partial \vec{c}} = \Delta \vec{w} \odot \frac{\partial L(\vec{u})}{\partial \vec{u}} \Bigg|_{\vec{u}=\vec{w} + \vec{c} \odot \vec{\Delta w}},
	\label{eq:efficient_importance_computation}
	\vspace{-0.05in}
	\end{equation}
	where $\odot$ denotes element-wise multiplication. When we set $\vec{c}=\vec{0}$, Eq. (\ref{eq:efficient_importance_computation}) becomes an importance vector where each element corresponds to a NN weight's importance. Since the loss gradient is parametrized by all the NN weights, the computed importance implicitly incorporates the impact of weight dependencies. 
	
	The importance of a tensor, then, is a summation of all its weights' importance, as described in Eq. (\ref{eq:importance}). To achieve additivity for such summation, our approach builds on the fact that the first-order components of the training loss function's Taylor expansion correspond to the importances of NN weights, as shown in Eq. (\ref{eq:expansion}) below:
	\begin{equation}
	L(\vec{w} + \Delta \vec{w}) = L(\vec{w}) + \frac{\partial L(w_1)}{\partial w_1} {\Delta w_1} + \frac{\partial L(w_2)}{\partial w_2} {\Delta w_2} + ...
	\label{eq:expansion}
	\end{equation}
	
	Since on-device training with a pre-trained model typically uses a very low learning rate (e.g., $10^{-4}$ to $10^{-5}$ as suggested in \cite{sharif2014cnn}) and values of weight updates ($\Delta w_i$) is proportional to the learning rate \cite{amari1993backpropagation}, the magnitudes of high-order components in the Taylor expansion are much lower than those of first-order components. Hence, we can use the first-order approximation, as described in Eq. (\ref{eq:expansion}), to retain additivity and calculate tensor importance.
	
	\begin{figure}[ht]
		\centering
		\vspace{-0.1in}
		\hspace{-0.25in}
		\subfigure[Different evaluation intervals] { 
			\includegraphics[width=0.24\textwidth]{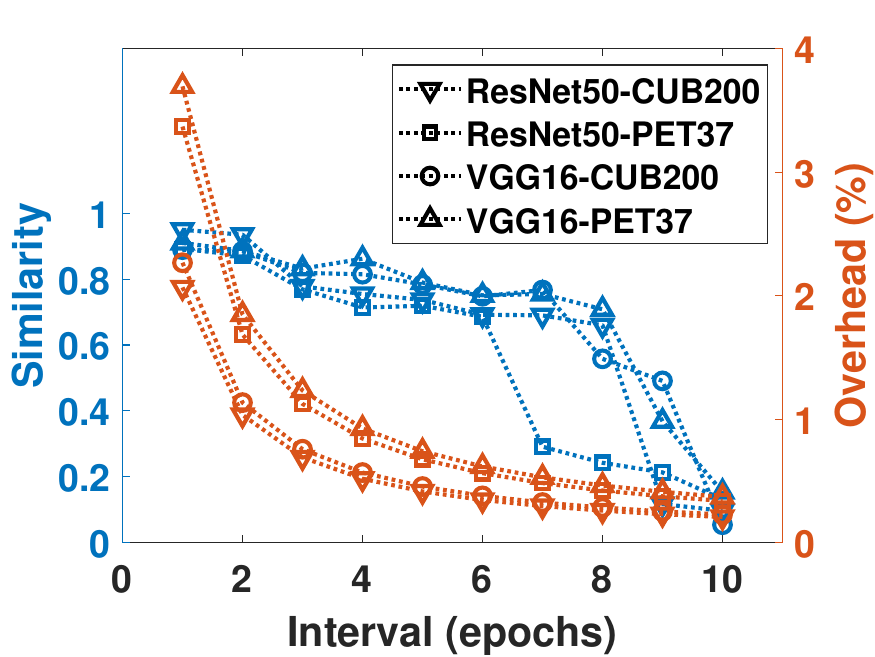}
			\label{fig:similarity_interval}
		}
		\subfigure[Different batch sizes] { 
			\includegraphics[width=0.24\textwidth]{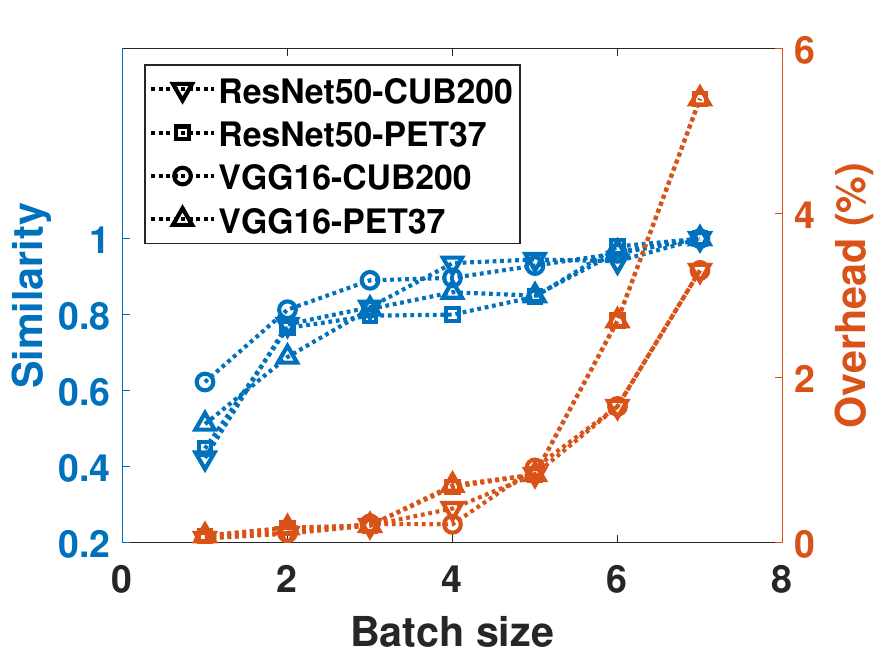}
			\label{fig:similarity_batchsize}
		}
		\hspace{-0.4in}
		\vspace{-0.15in}
		\caption{The impact of evaluation interval and batch size on tensor importance evaluation}
		\label{fig:performance_importance}
		\vspace{-0.2in}
	\end{figure}
	
	Importances of tensors need to be frequently evaluated at run-time during training. A long interval of evaluation could fail to capture the variations of tensor importances over time and hence reduce the evaluation accuracy, but too frequent evaluations incur extra computation overhead. To investigate the impact of such evaluation interval on training, we train the pre-trained ResNet50 and VGG16 models on the CUB-200 and PET-37 \cite{parkhi2012cats} datasets using Jetson TX2, and measure the cosine similarity between the NN tensor importances evaluated in every training epoch. Results in Figure \ref{fig:similarity_interval} show that such similarity generally reduces with longer intervals of importance evaluation, but can retain at 80\% when the interval is 3 training epochs. When such an interval is used, the computing overhead of importance evaluation is $<$1\% of the whole training cost, and we consider this value as the optimal for the interval of importance evaluation.
	
	Since the loss gradient is computed from a batch of uniformly sampled training data, the batch size also affects such similarity. As shown in Figure \ref{fig:similarity_batchsize}, using a batch size of 4 is sufficient to retain high similarity with $<$1\% of computation overhead. If needed, using a larger batch size is still possible by breaking the original batch into smaller micro-batches and aggregating their results.
	

	\vspace{-0.05in}
	\section{Profiling Execution Times of Tensor Training}
	\label{sec:profiling}
	To correctly profile how each selected tensor contributes to NN training time, we will explicitly find out how each tensor  associates with NN operations in the backward computing graph of different types of NN layers. Then, since timings of these NN operations can be obtained by standard NN profilers, we can estimate tensor's training time by matching and aggregating these operations' timings. 
	
	In practice, most modern AI frameworks (e.g., TensorFlow \cite{abadi2016tensorflow} and PyTorch \cite{paszke2019pytorch}) allow users to retrieve the list of NN trainable tensors sorted in their execution order (e.g., \texttt{model.trainable\_weights}). Based on such execution order, any complex NN model (e.g., with multiple branches) can be unrolled into a sequence of operations and the related tensors. We can then traverse this sequence to profile their execution times, and then perform necessary timing aggregation on these operations and tensors.
	
	\begin{figure}
		\centering
		\includegraphics[width=0.9\linewidth]{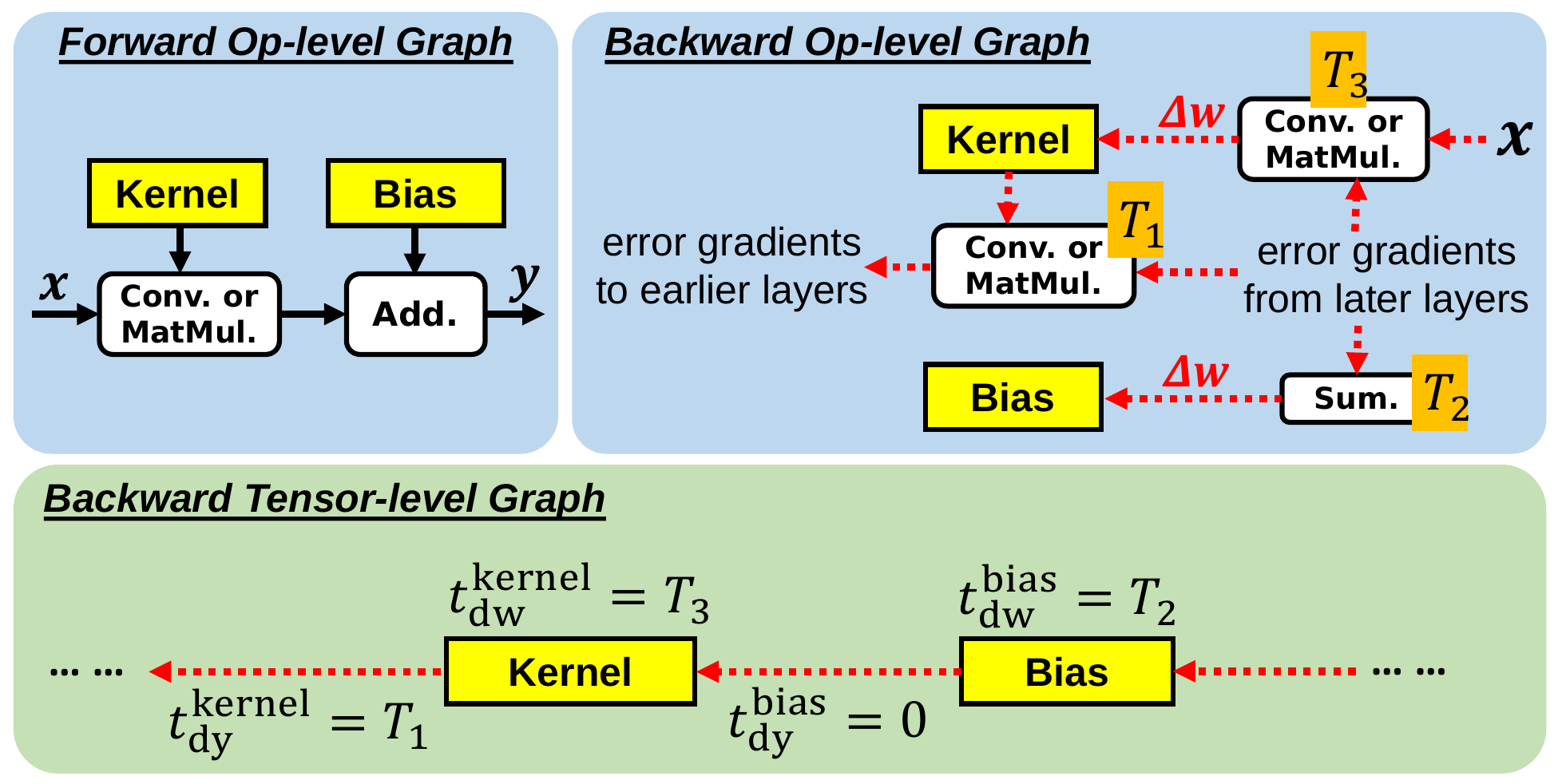}
		\vspace{-0.1in}
		\caption{Timing of tensor training in convolutional and dense layers}
		\vspace{-0.1in}
		\label{fig:conv_dense}
		\vspace{-0.05in}
	\end{figure}
	
	\subsection{Convolutional Layer \& Dense Layer}
	Both convolutional and dense layers\footnote{Dense layer is also known as Fully-connected layer.} contain two types of trainable tensors: \emph{Kernel} and \emph{Bias}. As shown in Figure \ref{fig:conv_dense}, in the forward pass, Kernel first multiplies and convolves the input data and the result is added by Bias to produce the layer output. In the backward pass, the error gradients from later layers are involved in three operations: 1) compute the updates of Kernel, 2) compute the updates of Bias\footnote{Since Bias is not involved in error gradient passing, its $t_{dy}$ equals to 0.}, and 3) pass the error gradient to the earlier layers, whose execution times are $T_3$, $T_2$, and $T_1$, respectively. As a result, we can convert the layer's computing graph from operation level to tensor level, where each tensor is equivalent to a standalone layer. The execution time of NN operations is then matched to the related tensors. This tensor-level computing graph exactly follows the time model of backward pass described in Section 2.2, and can be used to directly estimate the training time of any set of selected tensors.

	\begin{figure}[ht]
		\centering
		\vspace{-0.05in}
		\includegraphics[width=0.95\linewidth]{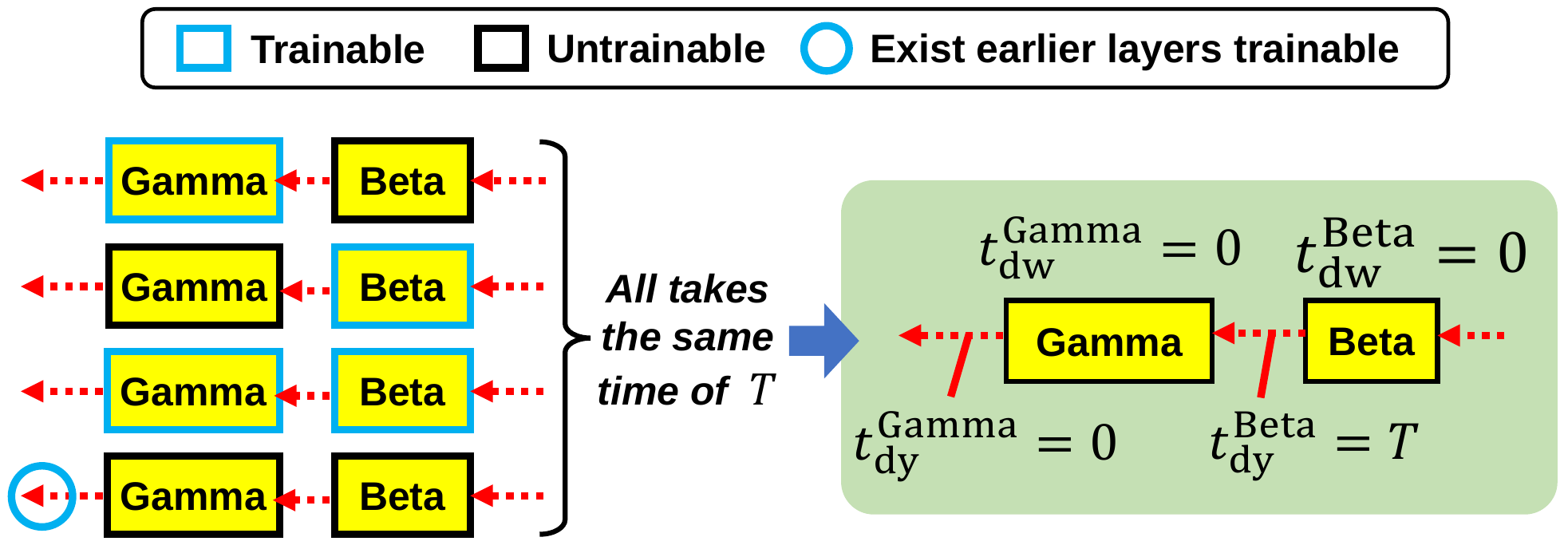}
		\vspace{-0.1in}
		\caption{Timing of tensor training in batch normalization layers}
		\vspace{-0.15in}
		\label{fig:batch_norm}		
	\end{figure}
	
	\vspace{-0.05in}
	\subsection{Batch Normalization Layer}
	A batch normalization (BN) layer \cite{ioffe2015batch} contains two types of trainable tensors: \emph{Gamma} and \emph{Beta}, and can be similarly profiled as described above. However, the implementations of BN layer may vary on different NN software frameworks that have different optimizations on normalization. To avoid the impact of this heterogeneity, we instead consider each BN layer as a black box and profile it by enumerating different ways of selecting its tensors in training. 
	
	As shown in Figure \ref{fig:batch_norm}, our experiments on widely used NN frameworks, including TensorFlow \cite{abadi2016tensorflow} and PyTorch \cite{paszke2019pytorch}, verify that as long as any tensor is selected in the current layer or any earlier layer, all the BN-related NN operations in these layers will be executed in the backward pass. This implies that, although BN has heterogeneous operation-level backward computing graphs, its tensor-level computing graph can be unified. Specifically, we aggregate all the BN-related operation time $T$ into $t_{dy}$ of the corresponding Beta. By doing so, as long as an error gradient is passed from Beta, the execution time of this tensor-level graph is $T$.

	
	\vspace{-0.05in}
	\subsection{Non-trainable Layer}
\begin{figure}[ht]
	\centering
	\vspace{-0.1in}
	\includegraphics[width=0.95\linewidth]{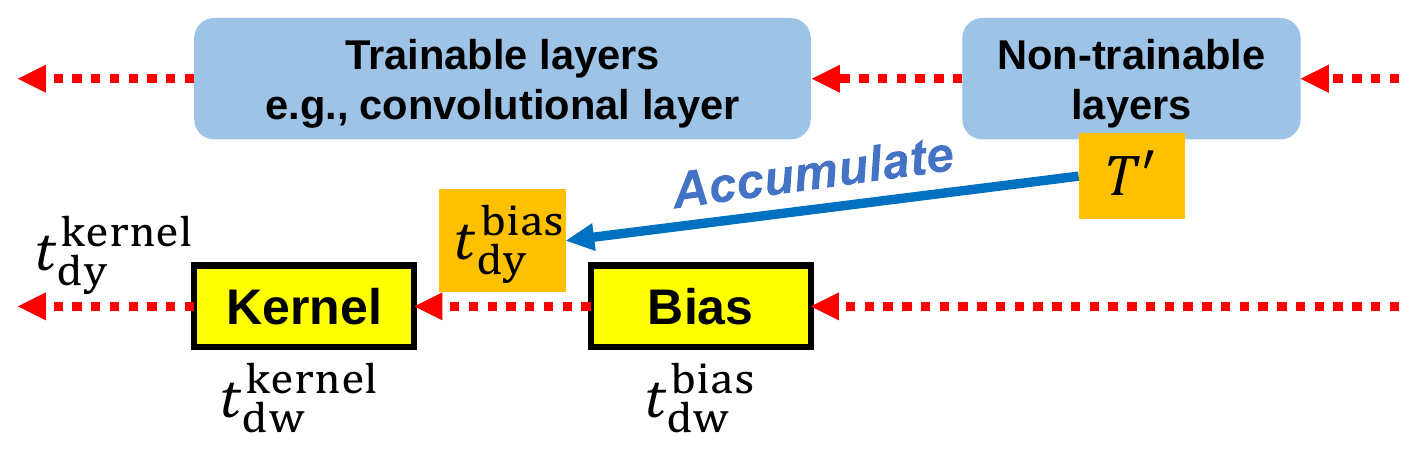}
	\vspace{-0.1in}
	\caption{Including timings of non-trainable layers}
	\label{fig:non_trainable}
	\vspace{-0.15in}
\end{figure}

	In addition, many other NN operations have no association with any trainable tensor. Instead, they belong to non-trainable layers such as activation and pooling layers. Although these layers have no tensor to update, they still incur computations to pass error gradients. To unify such behavior into the tensor-level computing graph, we incorporate the backward operation time of non-trainable layers into the timing of the closest tensor in the previous trainable layer. For example as shown in Figure \ref{fig:non_trainable}, the time $T'$ of all operations in a non-trainable layer is added to $t_{dy}$ of Bias in the previous convolutional layer. In this way, if the error gradient passes Bias, the timing of non-trainable layers will be guaranteed to be counted.

\subsection{Profiler's Reliability and Generality}
In practice, such offline profiling may not always capture the runtime variability of NN operations' timings. However, variation of such timings on embedded devices is usually small, because NN training is rarely co-executed with other computing tasks. On the other hand, when experiencing significant variations (e.g., due to throttling), it is feasible to frequently run the profiler online with very low cost. As shown in Table \ref{table:profile_overhead}, our experiments on different NN models show that profiling in one training epoch only incurs $<$1\% extra computing overhead.

\begin{table}[h]
	\vspace{-0.05in}
	\begin{tabular}{c||c|c|c|c}
		\hline
		NN model & ResNet50 & VGG16 & MobileNetV2 & ViT \\ \hline\hline
		Overhead & 0.42\%   & 1.00\%  & 0.60\% & 0.61\%  \\ \hline
	\end{tabular}
	\vspace{0.05in}
	\caption{Overhead of runtime profiling every epoch}
	\label{table:profile_overhead}
	\vspace{-0.3in}
\end{table}

Besides the types of NN layers discussed above, our proposed profiling approach is also generalizable to other NN architectures. Advanced NN components, by their designs, can be decomposed into a series of previously mentioned layers. For example, self-attention modules in Transformers \cite{vaswani2017attention} are constructed by mainly stacking dense layers (Section 5.1) and attention operations (Section 5.3). Other normalization layers (e.g., LayerNorm \cite{ba2016layer} and GroupNorm \cite{wu2018group}) share the same behavior with batch normalization (Section 5.2). Dropout layers \cite{baldi2013understanding} all belong to non-trainable layers (Section 5.3).

	\vspace{-0.05in}
	\section{Dynamic Programming for Tensor Selection}
	
	
	\subsection{Subproblem Definition}
\begin{figure}[h]
		\centering
		\vspace{-0.1in}
		\includegraphics[width=0.95\linewidth]{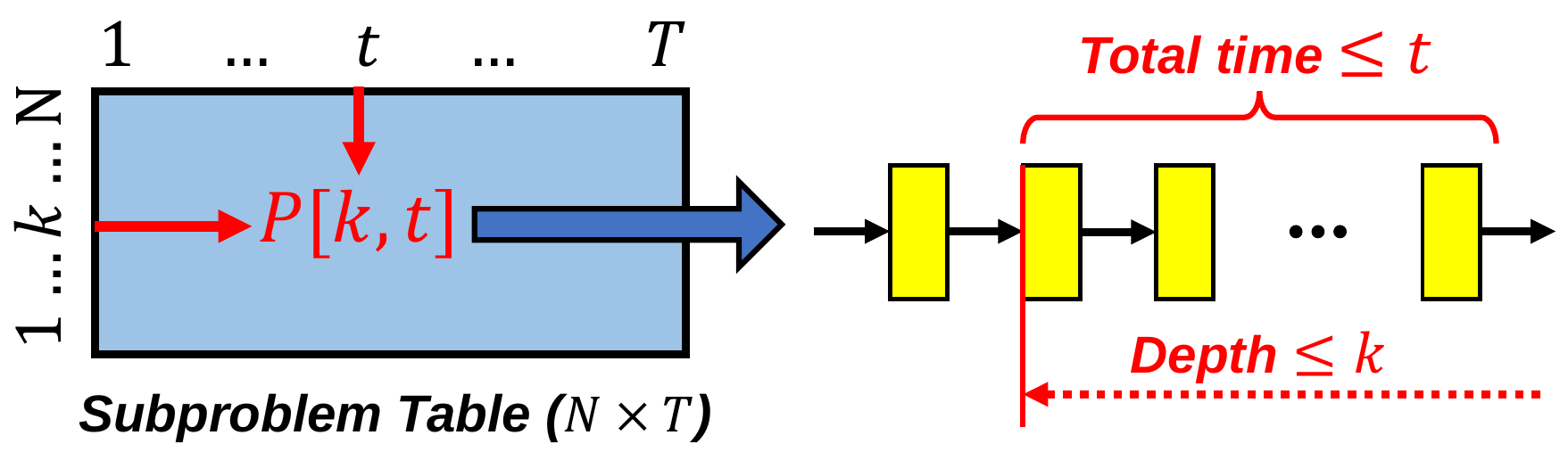}
		\vspace{-0.1in}
		\caption{Subproblem definition}
		\vspace{-0.05in}
		\label{fig:subproblem}
		\vspace{-0.05in}
	\end{figure}

	As shown in Figure \ref{fig:subproblem}, in our DP algorithm, we define each subproblem $P[k,t]$ as to maximize the cumulative importance of selected tensors when 1) selection is among the bottom $k$ tensors and 2) training time in backward pass is at most $t$. Since $t$ always appears as discrete values (e.g., in microseconds), its resolution is high enough to capture the disparity between different tensors' timings. 

	DP starts from solving the smallest subproblem with $t=1$ and $k=1$ and gradually solves larger subproblems based on the results of smaller subproblems. The key challenge is how to find the recurrence relation of these subproblems.

	\subsection{Recurrence Relation of Subproblems}
	Recurrence relation between subproblems $P[k,t]$ and $P[k-1,t]$ depends on whether we further select the bottom tensor $k$ from the solution of $P[k-1,t]$. As shown in Figure \ref{fig:recursion}:

	\begin{figure}[ht]
		\centering
		\vspace{-0.15in}
		\hspace{-0.25in}
		\subfigure[Bottom tensor $k$ is not selected] { 
			\includegraphics[width=0.24\textwidth]{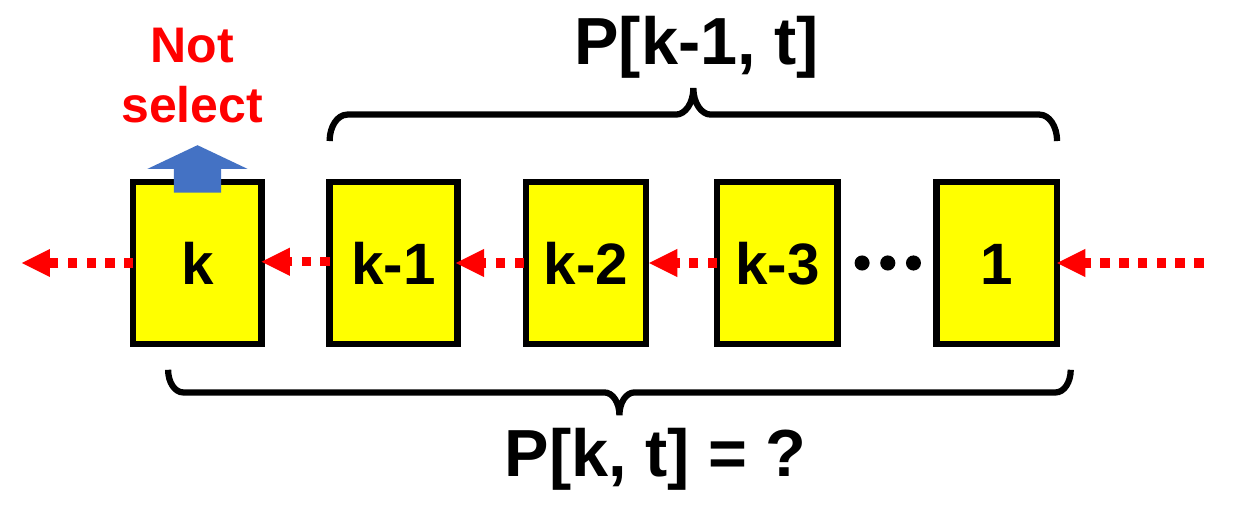}
			\label{fig:dp_case1}
		}
		\subfigure[Bottom tensor $k$ is selected] { 
			\includegraphics[width=0.24\textwidth]{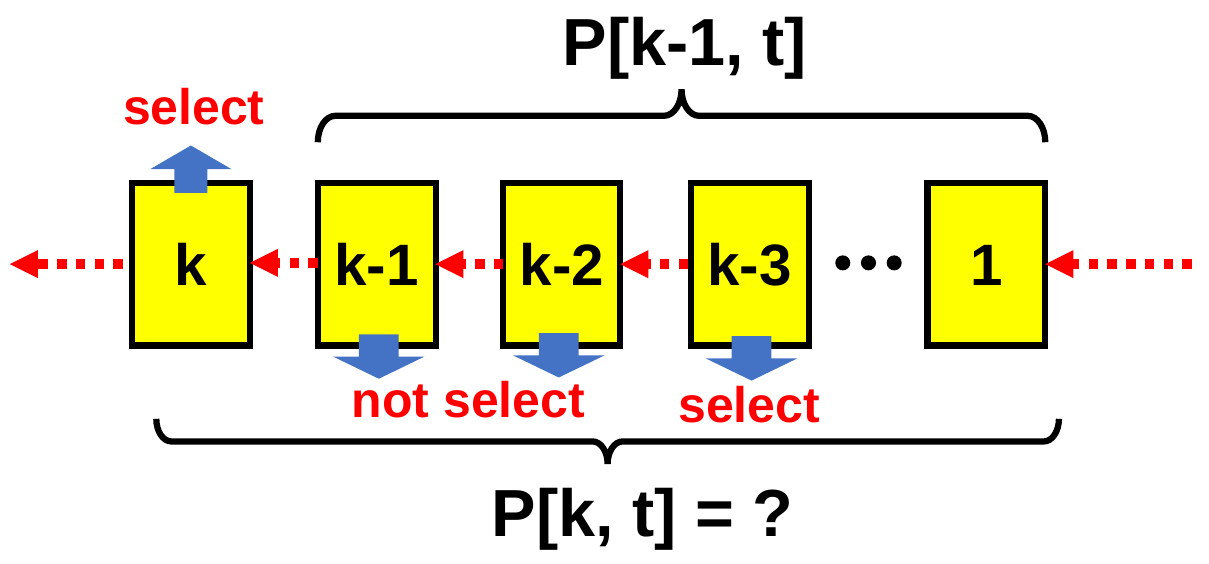}
			\label{fig:dp_case2}
		}
		\hspace{-0.4in}
		\vspace{-0.15in}
		\caption{Finding recursion relation in different cases}
		\label{fig:recursion}
		\vspace{-0.1in}
	\end{figure}
	
	\noindent\textbf{Case 1:} If the bottom tensor $k$ is not selected, $P[k,t]$ will fall back to $P[k-1,t]$ since the importance of tensor selection cannot be further increased. 
	
	\noindent\textbf{Case 2:} If the bottom tensor $k$ is selected, then two timings will be surely included in the solution of $P[k,t]$, no matter which other tensors are selected: 1) the time to update tensor $k$; 2) the time to pass error gradient from the closest selected tensor $k_c$, such as tensor $k-3$ as shown in Figure \ref{fig:dp_case2}, to tensor $k$. This implies that $P[k,t]$ can fall back to a previously solved subproblem $P[k-k_c,t-\Delta t]$ where:
	\begin{equation} 
	\Delta t = t_{dw}^{(k)}+\sum\nolimits_{j=k_c}^{k-1}{t_{dy}^{(j)}}.
	\label{eq:recursion}
	\end{equation}
	
	Since $k_c$ is unknown in advance, we backtrace the previously solved subproblems and explore all the possibilities of $k_c$, by reducing the depth of backward pass from $k$. 
	
	As a result, the optimal solution to $P[k,t]$ is the one with higher cumulative importance of selected tensors between Case 1 and 2. Based on this recurrence relation, we can solve all the subproblems by sequentially traversing the subproblem space. The time complexity of solving each subproblem is $O(N)$ due to backtracing in Case 2, and the overall time complexity of DP algorithm is $O(N^2T)$.

	
	
	\begin{figure}[ht]
		\centering
		\vspace{-0.2in}
		\hspace{-0.25in}
		\subfigure[Skipping subproblems] { 
			\includegraphics[width=0.24\textwidth]{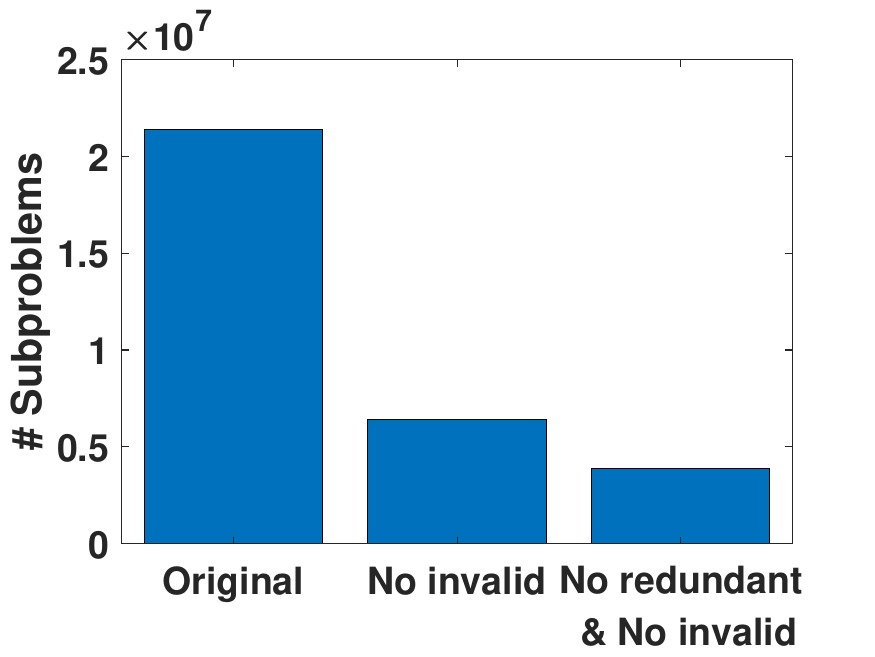}
			\label{fig:skip_subproblems}
		}
		\subfigure[Reducing the time resolution] { 
			\includegraphics[width=0.24\textwidth]{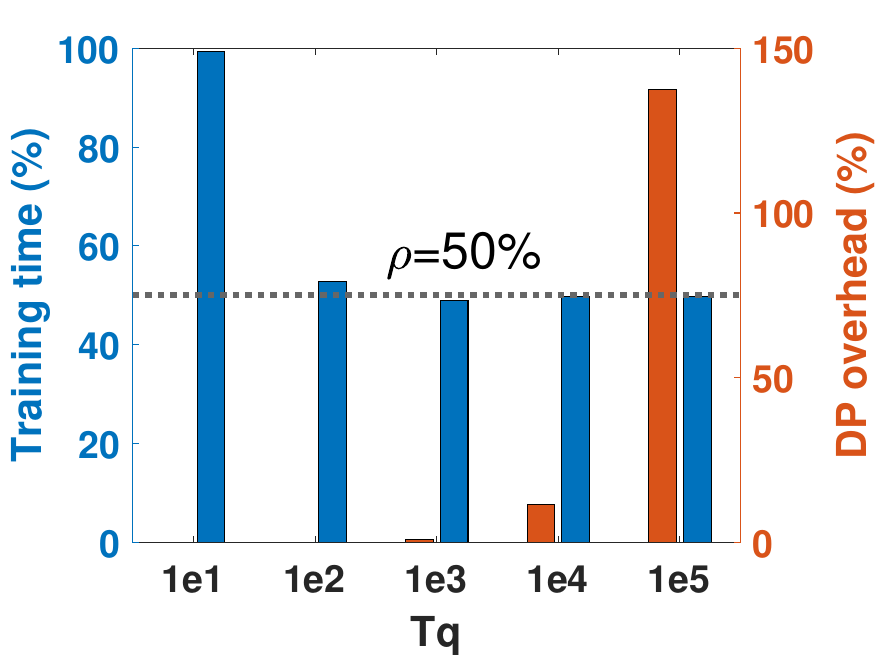}
			\label{fig:dp_scale}
		}
		\hspace{-0.4in}
		\vspace{-0.15in}
		\caption{Reducing the computing cost of DP}
		\label{fig:reduce_dp_time}
		\vspace{-0.15in}
	\end{figure}
	
	\subsection{Reducing the Computational Cost}
	
	Due to the large value of training time in wall-clock time units, there could be $>$100M subproblems in practical scenarios. To reduce the computational cost of DP, we reduce the subproblem space by skipping two types of subproblems: 1) \emph{invalid ones}, whose time constraint $t$ plus the forward pass time exceeds the desired timing constraint ($\rho T$); 2) \emph{redundant ones}, whose time to pass error gradient to the maximally allowed depth ($k$) exceeds $t$. As shown in Figure \ref{fig:skip_subproblems}, doing so on the ResNet50 model with $\rho=50\%$ can reduce the number of subproblems by 5.5$\times$ without affecting optimality.
	
	To further reduce the number of subproblems, we scale tensors' timings ${t_{dw}}$ and ${t_{dy}}$ by multiplying a factor of $Z$:
	\begin{equation} 
	\widetilde{t_{dw}} = \left\lfloor t_{dw}\cdot{Z} \right\rfloor, \ \ \ \ \widetilde{t_{dy}} = \left\lfloor t_{dy}\cdot{Z} \right\rfloor, \ \ \ \  
	\label{eq:downscaling}
	\end{equation}
	where $Z = \frac{T_q}{T}$ and the backward pass time is reduced to a resolution $T_q<T$. The overall time complexity of DP is then reduced to $O(N^2T_q)$. Such reduced resolution could increase the ambiguity in DP and affect optimality. To avoid this issue, we will run DP with different resolutions and find the best resolution that balances optimality and computational cost. For example, for a ResNet50 model trained with CUB-200 dataset and $\rho=50\%$, Figure \ref{fig:dp_scale} shows that $T_q=10^3$ can reduce the DP execution time to $<1$\% of training time but still achieve the desired training speedup. 
	
	On the other hand, since the time complexity of DP increases quadratically with $N$, its computing cost could still be high when being applied to very large NN models (e.g., GPT-3 \cite{brown2020language}). In these cases, we could further leverage the existing parallelization techniques (e.g., multithreading \cite{tan2008improving}) and hardware accelerators (e.g., GPUs) to speed up DP.

	\begin{figure}[ht]
		\centering
		\vspace{-0.1in}
		\includegraphics[width=0.6\linewidth]{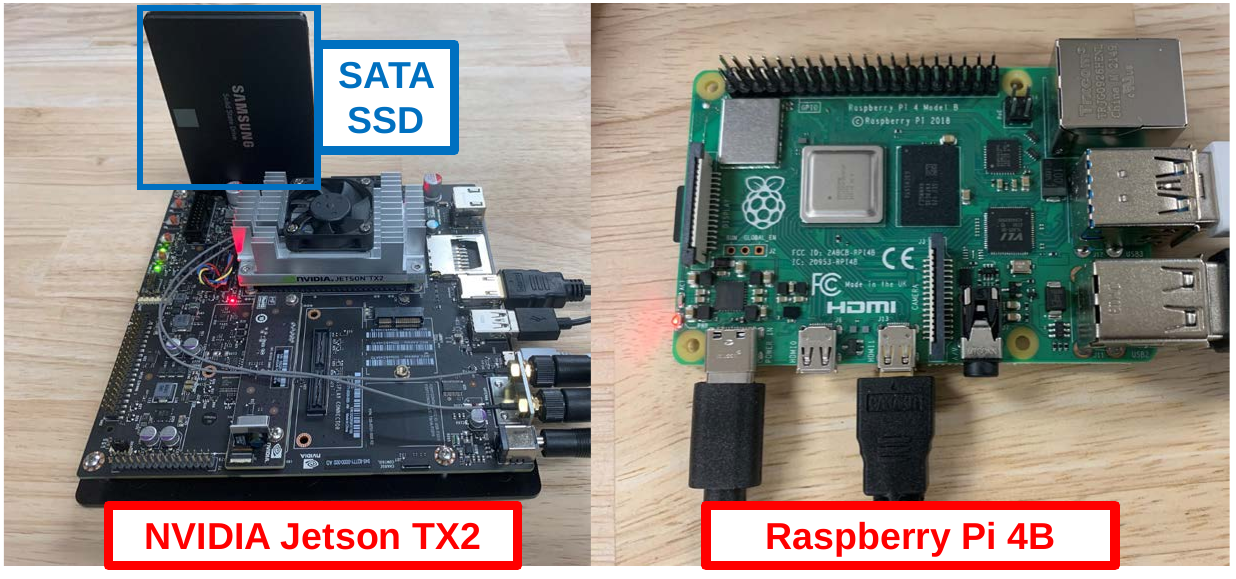}
		\vspace{-0.1in}
		\caption{Devices used in our implementation}
		\vspace{-0.15in}
		\label{fig:device_setup}
	\end{figure}
	
	\section{Implementation}
	As shown in Fig \ref{fig:device_setup}, we implement ElasticTrainer on embedded devices including Nvidia Jetson TX2 and Raspberry Pi 4B, both of which are widely used AI computing devices on embedded platforms such as drones \cite{skydio} and small robots \cite{jetbot}. Jetson TX2 is equipped with a Nvidia Pascal GPU with 256 CUDA cores, a 1.2GHz Cortex-A57 CPU and 8GB memory shared between GPU and CPU. Raspberry Pi 4B is equipped with a 1.5GHz Cortex-A72 CPU and 4GB memory. We install Jetpack 4.6.2 OS based on Ubuntu 18.04 on Jetson TX2, and 64-bit Raspbian 11 OS on Raspberry Pi.
	
	We conduct training using TensorFlow 2.7 Python API and TensorFlow Addons 0.15, which are compiled from TensorFlow source codes on both devices. We use TensorFlow's profiler plugin to profile the execution time of NN operations. Due to TensorFlow's limited support\footnote{https://github.com/tensorflow/tensorflow/issues/56434}, using GPUs for training on ARM-based platforms could cause gradual memory leakage over time. To prevent potential core dump, as shown in Figure \ref{fig:device_setup}, we attach an external SATA SSD to Jetson TX2 to increase its swap space in training. On the other hand, we interact with Jetson TX2 using the text-only interface, to reduce interference from the graphics of other apps that could affect the evaluation results. Similarly, we use the text-only interface on Raspberry Pi 4B, where the local memory is sufficient because the memory leakage only exhibits significant accumulation over time on GPU devices.
	
	\begin{figure*}
		\centering
		\hspace{0.35in}
		\subfigure[Accuracy on CUB-200] { 
			\includegraphics[width=0.26\textwidth]{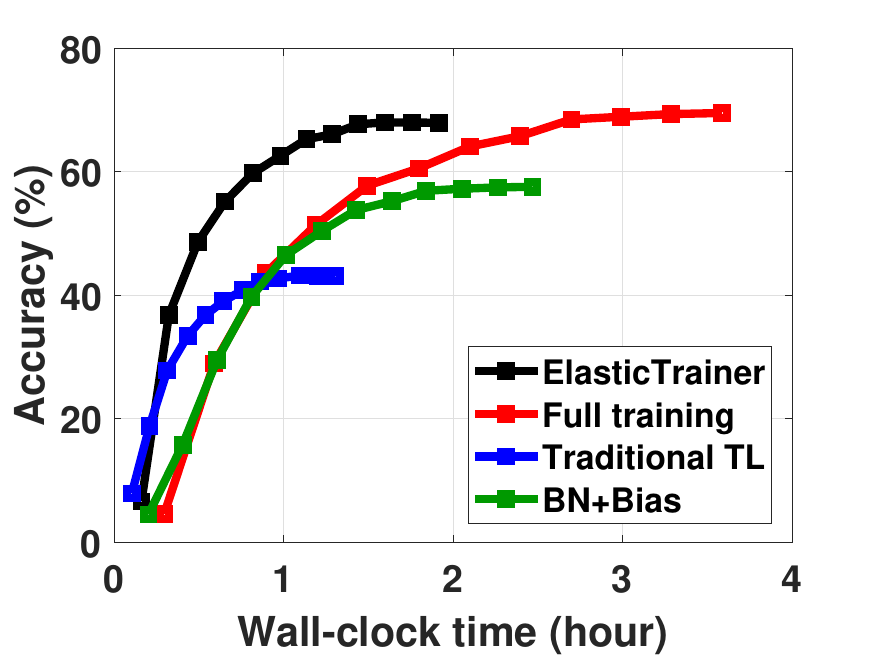}
			\label{fig:resnet50_cub200_acc_time_curve}
		}
		\hspace{0.2in}
		\subfigure[Accuracy on Oxford-IIIT Pet] { 
			\includegraphics[width=0.26\textwidth]{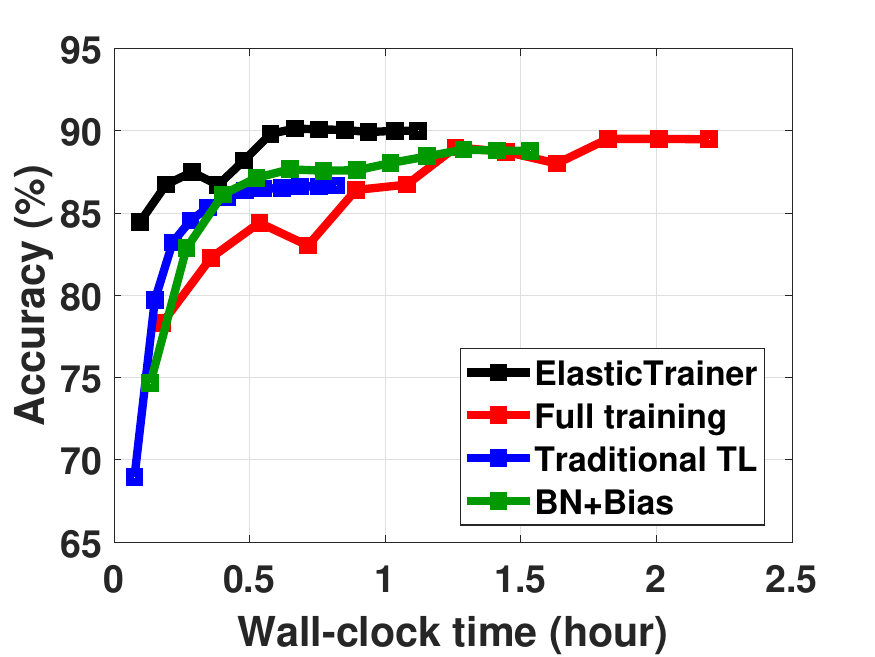}
			\label{fig:resnet50_pet37_acc_time_curve}
		}
		\hspace{0.2in}
		\subfigure[Accuracy on Stanford Dogs] { 
			\includegraphics[width=0.26\textwidth]{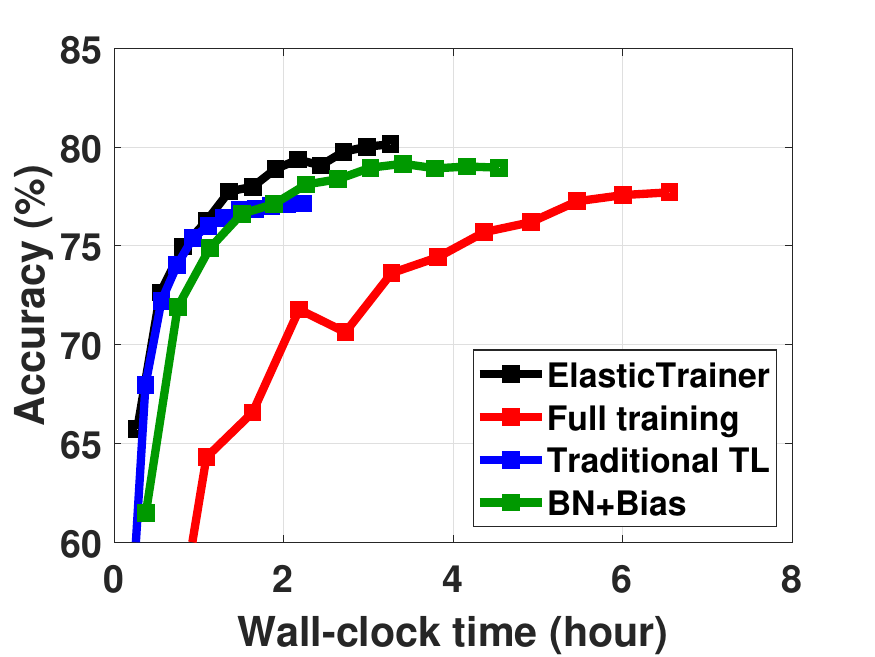}
			\label{fig:resnet50_dogs120_acc_time_curve}
		}
		\hspace{0.2in}
		\newline
		\vspace{-0.2in}
		\centering
		\subfigure[Loss on CUB-200] { 
			\includegraphics[width=0.26\textwidth]{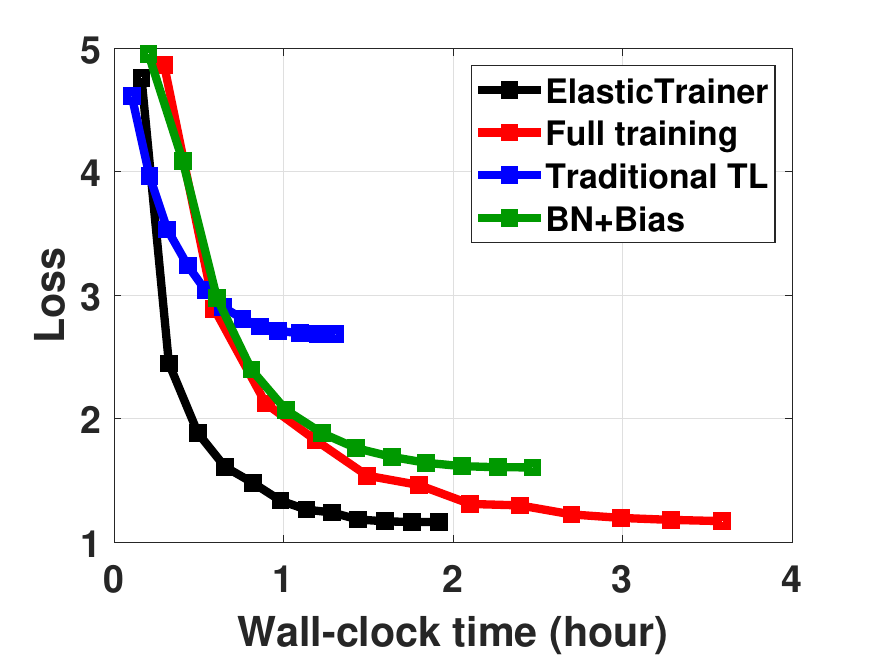}
			\label{fig:resnet50_cub200_loss_time_curve}
		}
		\hspace{0.2in}
		\subfigure[Loss on Oxford-IIIT Pet] { 
			\includegraphics[width=0.26\textwidth]{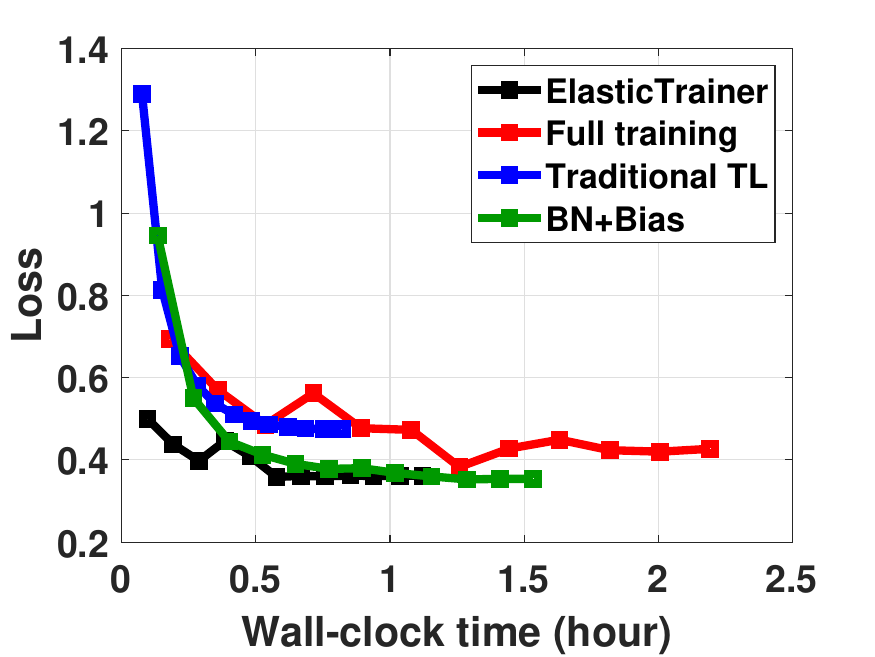}
			\label{fig:resnet50_pet37_loss_time_curve}
		}
		\hspace{0.2in}
		\subfigure[Loss on Stanford Dogs] { 
			\includegraphics[width=0.26\textwidth]{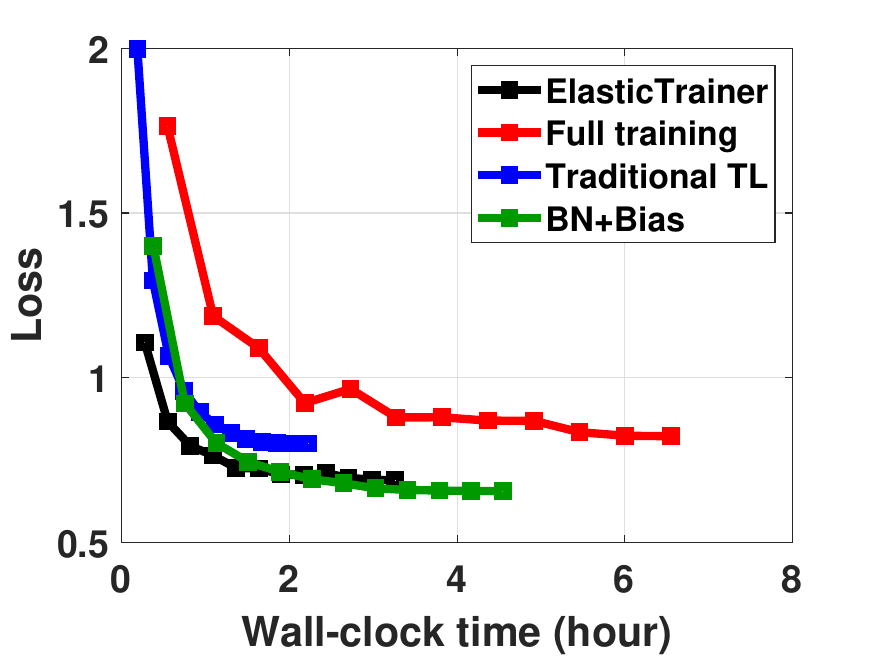}
			\label{fig:resnet50_dogs120_loss_time_curve}
		}
		
		\caption{Testing accuracy and training loss over time with different datasets on Jetson TX2}
		\label{fig:jetson_curves}
		\vspace{-0.1in}
	\end{figure*}

	\vspace{-0.05in}
	\section{Performance Evaluation}
	In our evaluation, we choose widely used NN models: ResNet50 \cite{he2016deep}, VGG16 \cite{simonyan2014very}, MobileNetV2 \cite{sandler2018mobilenetv2}, and Vision Transformer (ViT) \cite{dosovitskiy2020image}, which are pre-trained using the ImageNet dataset \cite{deng2009imagenet} with 1.2M images. We then use the following new and fine-grained datasets with more complicated data patterns and fewer training samples in each category (e.g., 20-100 samples per class) to perform on-device training on the pre-trained models:
	\begin{itemize}
		\item \textbf{CUB-200 \cite{WahCUB_200_2011}} with 5,994 training and 5,794 testing images from 200 bird species for classification. Only about 20 training samples are provided for each species, which has $>$20 diverse attributes such as crown color and wing shape. All the photos are taken from the wild with different scales and highly variant backgrounds. 
		\item \textbf{Oxford-IIIT Pet \cite{parkhi2012cats}} with 3,680 training and 3,669 testing images from 37 categories of pets. The images have large variations in scale, pose and lighting to incorporate real-world random factors.
		\item \textbf{Stanford Dogs \cite{khosla2011novel}} with 12,000 training and 8,580 testing images of 120 dog breeds. Images in each breed have dogs in different ages, poses, and colors.
	\end{itemize}
	
	In this way, our evaluations target difficult learning tasks, compared to those in the existing work \cite{zhang2020mdldroidlite,li2021hermes} that only involves simple datasets such as MNIST \cite{deng2012mnist} and CIFAR-10 \cite{krizhevsky2009learning} with easy patterns and sufficient training samples (e.g., $>$5,000 per class). Since the NN's output dimension varies with different datasets, we replace the original output layer in the pre-trained model with randomly initialized parameters to fit each dataset's requirement before on-device training. We compare the performance of ElasticTrainer with the following four baselines:
	
	
	\begin{itemize}
		\item \textbf{Full training:} All parameters in the pre-trained model are trained on the new dataset. 
		\item \textbf{Traditional TL \cite{donahue2014decaf, sharif2014cnn}:} Only parameters in the last prediction module are continually trained. This is widely adopted in traditional transfer learning for low cost and simplicity.
		\item \textbf{BN+Bias \cite{cai2020tinytl, mudrakarta2018k}:} Besides the last prediction module, bias parameters in convolutional and dense layers, and parameters in batch normalization layers are involved in offline selection for on-device training. These two types of parameters have been identified to be important for efficient training.
		\item \textbf{PruneTrain \cite{lym2019prunetrain}:} The NN is trained with runtime pruning. It gradually reduces NN training cost by pruning less important channels in convolutional layers.
	\end{itemize}
		
The actual performance of convergence in NN model training would largely depend on the hyper-parameter settings, such as the batch size and learning rate. For fair comparisons, being similar to the existing work \cite{zhang2020mdldroidlite, cai2020tinytl}, we restrict the training of all NN models to 12 epochs as the required training time, which also corresponds to our objective of the training speedup. No matter whether the training converges or not within 12 epochs, we aim to maximize the model's prediction accuracy that can be achieved within this time frame.
	
	In all experiments, we use batch size of 4 and resize images into 224$\times$224\footnote{Using a higher resolution (e.g., 448$\times$448) and a larger batch size (e.g., 64) may lead to better model accuracy, but is very memory consuming and hence not applicable on typical embedded devices such as Raspberry Pi and Nvidia Jetson.} and apply default pre-processing steps such as pixel centering and random flipping before training. Such resizing and pre-processing are executed on the target device at runtime, and their computing overheads are counted as part of on-device training time. For training ViT, we use the Adam optimizer \cite{kingma2014adam} with a learning rate of $1\times10^{-4}$ without decay and scheduling. For the other models, we use the SGDW optimizer \cite{loshchilov2017decoupled} with a learning rate of $1\times10^{-4}$, momentum of 0.9, and weight decay of $5\times10^{-4}$. The learning rate is scheduled with standard cosine decay in training.


	\vspace{-0.05in}
	\subsection{Training Speedup \& Accuracy}
	We first compare ElasticTrainer with other baseline schemes on 3 different datasets. We use the ResNet50 NN model and set the speedup ratio $\rho$ to 50\%, which indicates an objective of achieving 2x training speedup. To better demonstrate the training progress, we record NN's testing accuracy and loss for every epoch. The computing cost on testing is also counted into the training time.

	\begin{figure}[ht]
		\centering
		\vspace{-0.1in}
		\hspace{-0.15in}
		\subfigure[Accuracy on CUB-200] { 
			\includegraphics[width=0.24\textwidth]{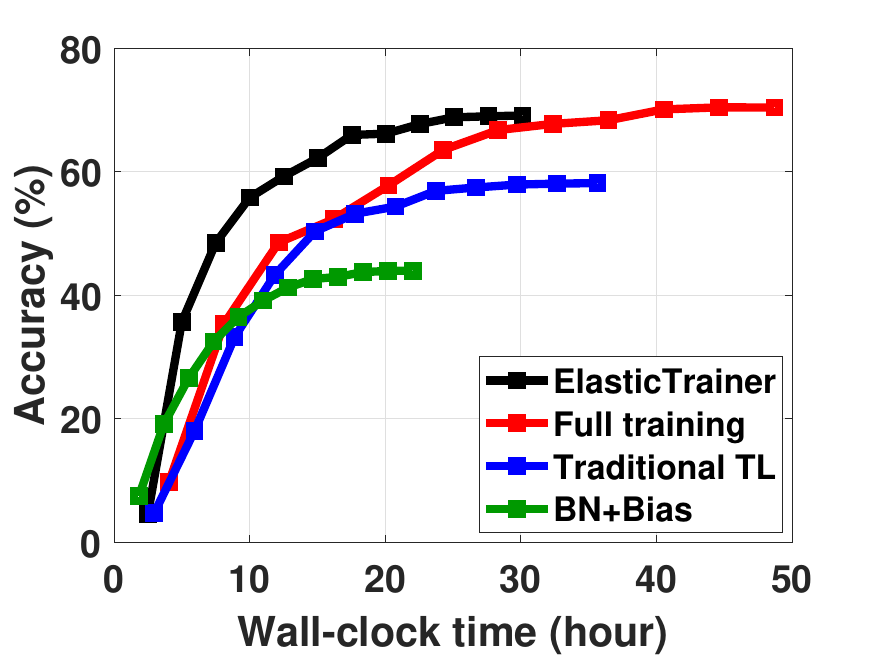}
			\label{fig:rpi_resnet50_cub200_acc_time_curve}
		}
		\hspace{-0.15in}
		\subfigure[Accuracy on Oxford-IIIT Pet] { 
			\includegraphics[width=0.24\textwidth]{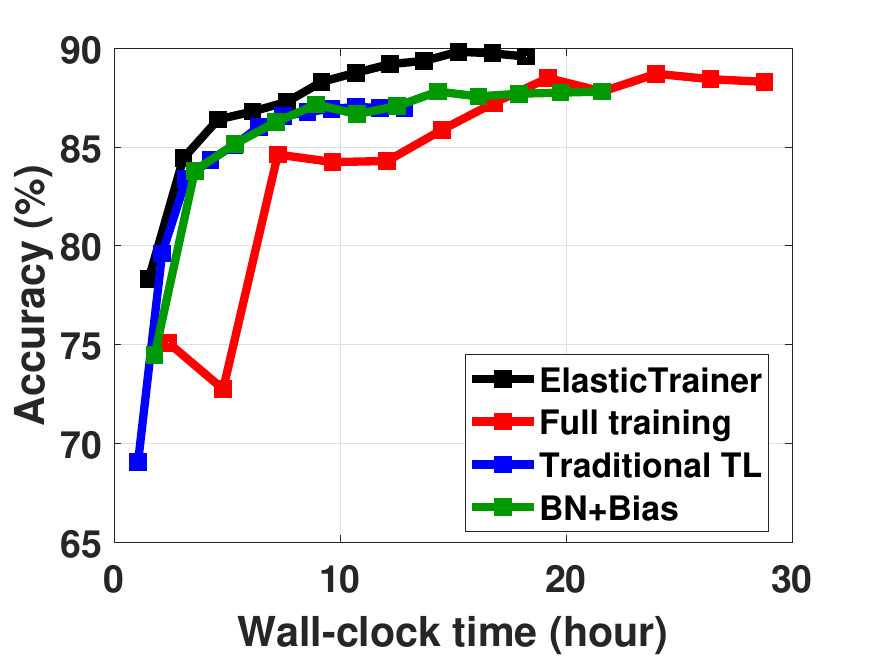}
			\label{fig:rpi_resnet50_pet37_acc_time_curve}
		}
		\hspace{-0.4in}
		\newline
		\vspace{-0.2in}
		\hspace{-0.25in}
		\subfigure[Loss on CUB-200] { 
			\includegraphics[width=0.24\textwidth]{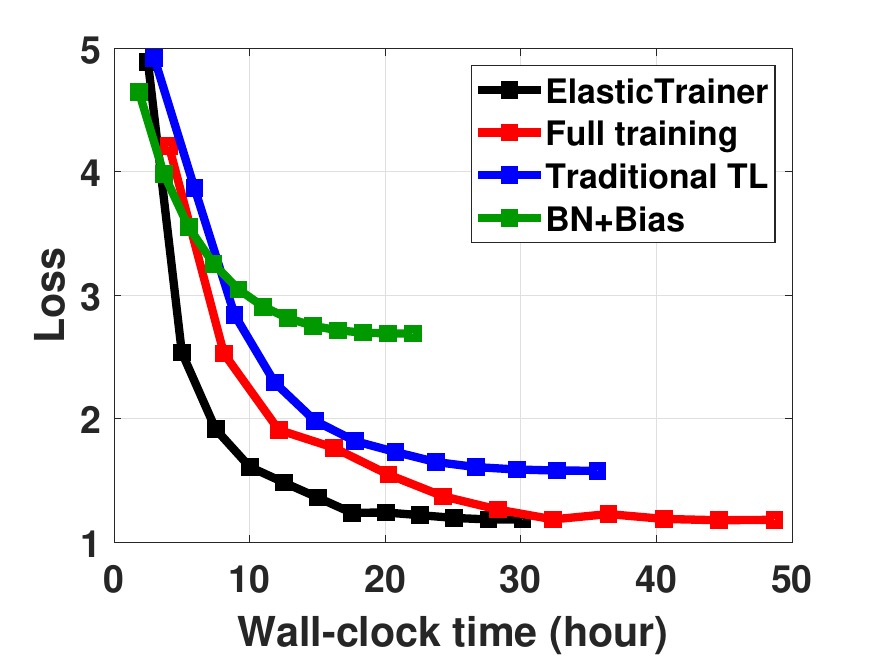}
			\label{fig:rpi_resnet50_cub200_loss_time_curve}
		}
		\hspace{-0.15in}
		\subfigure[Loss on Oxford-IIIT Pet] { 
			\includegraphics[width=0.24\textwidth]{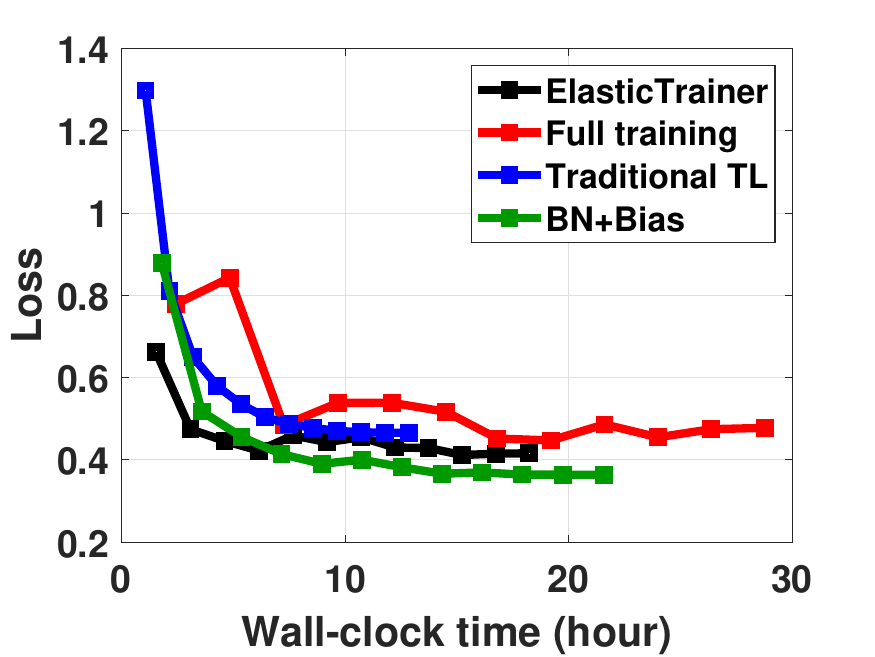}
			\label{fig:rpi_resnet50_pet37_loss_time_curve}
		}	
		\hspace{-0.4in}
		\caption{Testing accuracy and training loss over time with different datasets on Raspberry Pi 4B}
		\label{fig:pi_curves}
		\vspace{-0.2in}
	\end{figure}

	Results in Figure \ref{fig:jetson_curves} and \ref{fig:pi_curves} show that, ElasticTrainer can achieve similar testing accuracy when compared to Full Training, even with a much smaller trainable NN portion. In particular, on simpler datasets such as Oxford-IIIT Pet and Stanford Dogs, ElasticTrainer even achieves 1\%-2\% higher accuracy due to less overfitting. Comparatively, Traditional TL loses $>$20\% accuracy on CUB-200 dataset compared to Full Training, due to its insufficient representational power of trainable NN portion. BN+Bias shows similar performance with ElasticTrainer on Stanford Dogs dataset due to less variant patterns between offline and online data, but its optimality significantly degrades on more difficult datasets such as CUB-200, with $>$10\% accuracy drop.
	

With a smaller trainable NN portion, ElasticTrainer can achieve 2$\times$-3.4$\times$ speedup on Jetson TX2 compared to full training. On all datasets, its training time is within 2.7 hours, which matches the average idle time available for on-device training per day \cite{xu2018deeptype}. Its training speedup also outperforms that of BN+Bias by up to 30\%. On the other hand, although Traditional TL provides 20\% extra training speedup, it exhibits the highest training accuracy reduction. Similar advantages of training speedup compared to full training are also demonstrated in Figure \ref{fig:pi_curves} on Raspberry Pi 4B. BN+Bias achieves 25\% more training speedup on Raspberry Pi 4B, at the cost of 30\% accuracy reduction on difficult datasets.
	
	\begin{table}[ht]
		\vspace{-0.05in}
		\begin{tabular}{cc||cc||cc}
			\hline
			\multicolumn{2}{c||}{\textbf{CUB-200}}                 & \multicolumn{2}{c||}{\textbf{Oxford-IIT Pet}}          & \multicolumn{2}{c}{\textbf{Stanford Dogs}}           \\ \hline 
			\hline
			\multicolumn{1}{c|}{Accu.} & Speedup      & \multicolumn{1}{c|}{Accu.} & Speedup      & \multicolumn{1}{c|}{Accu.} & Speedup      \\ \hline 
			\multicolumn{1}{c|}{60\%}     & 2.19$\times$ & \multicolumn{1}{c|}{85\%}     & 2.94$\times$ & \multicolumn{1}{c|}{77\%}     & 1.80$\times$ \\
			\multicolumn{1}{c|}{64\%}     & 1.86$\times$ & \multicolumn{1}{c|}{87\%}     & 2.52$\times$ & \multicolumn{1}{c|}{78\%}     & 1.67$\times$ \\
			\multicolumn{1}{c|}{68\%}     & 1.88$\times$ & \multicolumn{1}{c|}{89\%}     & 2.44$\times$ & \multicolumn{1}{c|}{79\%}     & 1.79$\times$ \\ \hline
		\end{tabular}
		\vspace{0.05in}
		\caption{Speedup vs. the best baseline on Jetson TX2}
		\vspace{-0.25in}
		\label{table:time_acc_efficiency}
	\end{table}

	In some cases, the plateau of testing accuracy or training loss by the end of 12 epochs may not always indicate convergence \cite{he2016deep}. Alternatively, training can also be stopped early when reaching a specific accuracy target, and the corresponding training speedup is then interpreted as time-to-accuracy efficiency as suggested by the existing work \cite{zhang2020mdldroidlite, li2022pyramidfl}. As shown in Table \ref{table:time_acc_efficiency}, with different accuracy targets, ElasticTrainer achieves extra training speedup by 1.67$\times$-2.94$\times$ compared to the existing schemes on different datasets.
	

	\begin{figure}[ht]
		\centering
		\vspace{-0.15in}
		\hspace{-0.15in}
		\subfigure[Time on Jetson TX2] { 
			\includegraphics[width=0.23\textwidth]{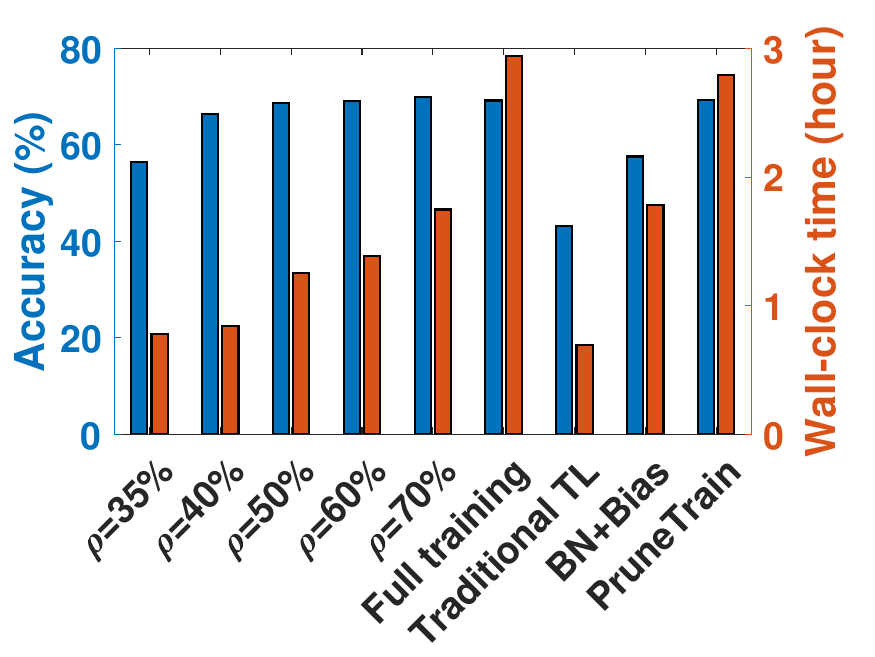}
			\label{fig:resnet50_acc_time_vs_all_jetson}
		}
		\hspace{-0.1in}
		\subfigure[FLOPs on Jetson TX2] { 
			\includegraphics[width=0.23\textwidth]{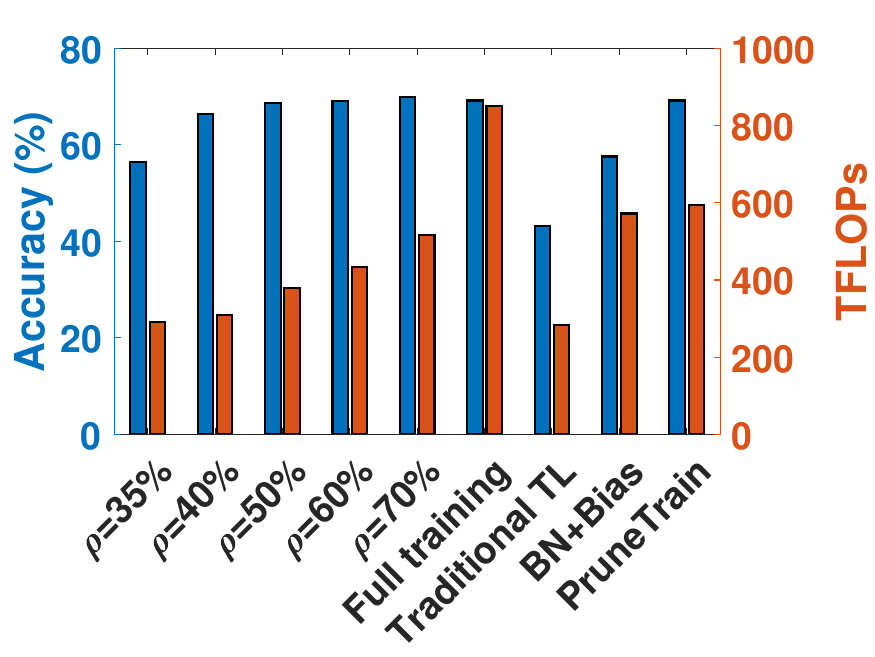}
			\label{fig:resnet50_acc_flops_vs_all_jetson}
		}\hspace{-0.4in}
		\newline
		\vspace{-0.2in}
		\hspace{-0.25in}
		\subfigure[Time on Raspberry Pi] { 
			\includegraphics[width=0.23\textwidth]{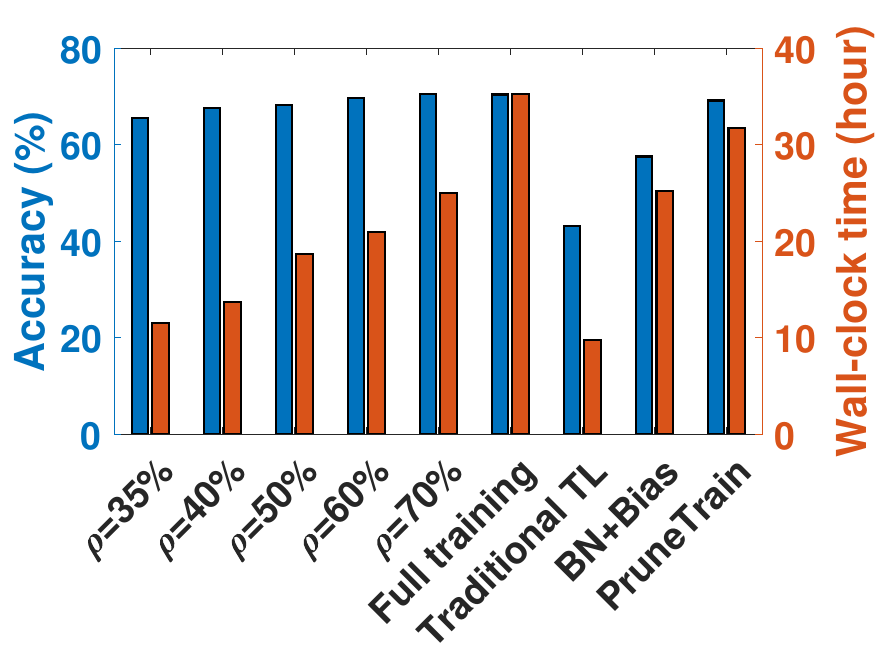}
			\label{fig:resnet50_acc_time_vs_all_pi}
		}
		\hspace{-0.1in}
		\subfigure[FLOPs on Raspberry Pi] { 
			\includegraphics[width=0.23\textwidth]{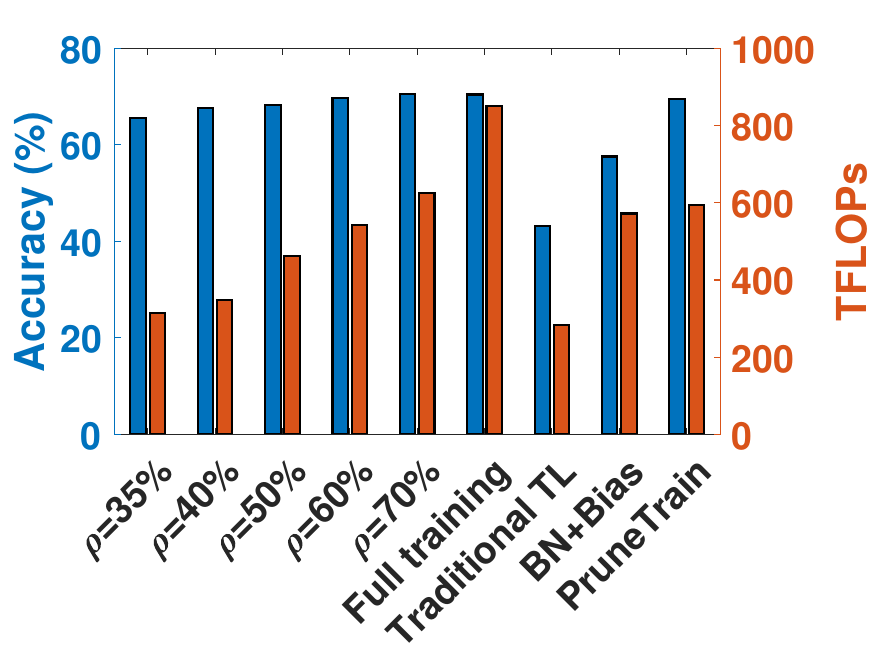}
			\label{fig:resnet50_acc_flops_vs_all_pi}
		}
		\hspace{-0.4in}
		\caption{Training Time \& FLOPs on CUB-200 with different $\rho$ on Jetson TX2 and Raspberry Pi 4B}
		\label{fig:different_rho}
		\vspace{-0.1in}
	\end{figure}
	    
	\vspace{-0.05in}
	\subsection{Impact of Speedup Objective}
	Different objectives of training speedups can be applied to on-device training, with different values of $\rho$. A lower value of $\rho$ asks for a smaller trainable NN portion and could hence achieve more training speedup, but may also reduce the NN accuracy due to insufficient NN representation power.

	\begin{figure}[ht]
		\centering
		\vspace{-0.1in}
		\includegraphics[width=0.85\linewidth]{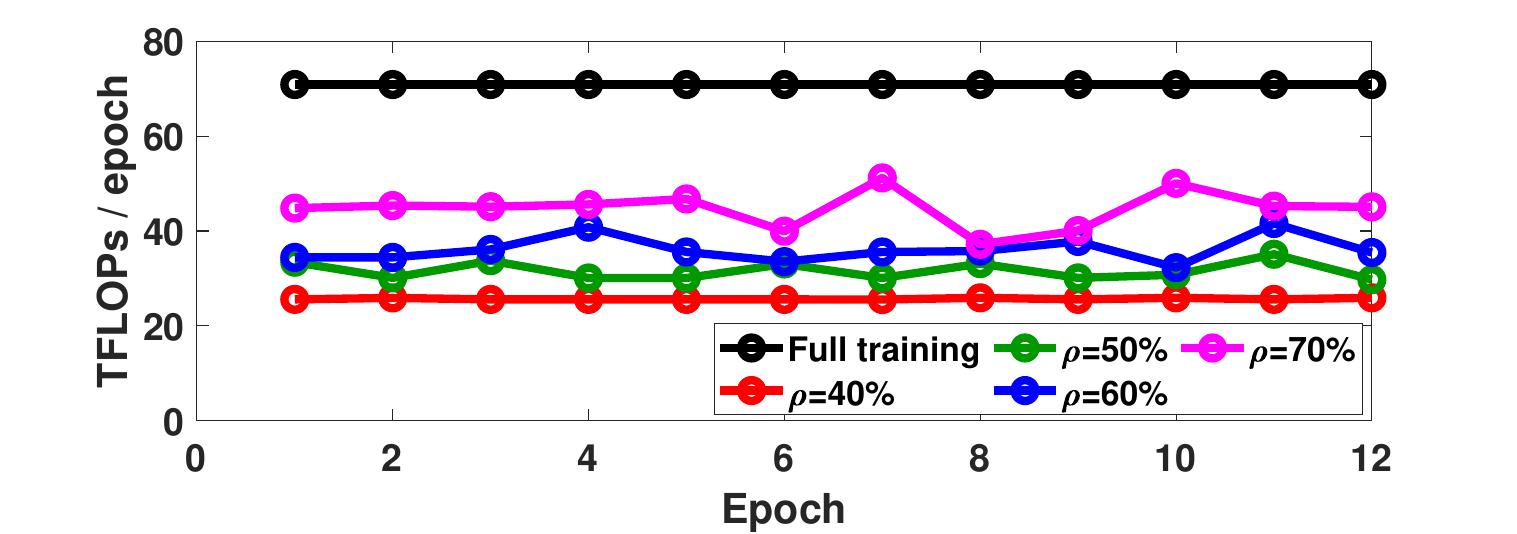}
		\vspace{-0.1in}
		\caption{FLOPs at every training epoch on CUB-200 with different $\rho$ on Jetson TX2}
		\vspace{-0.1in}
		\label{fig:flops_vs_epoch}
	\end{figure}

	To evaluate the impact of such speedup objective, we use the ResNet50 NN model and vary the value of $\rho$ from 35\% to 70\%.  As shown in Figure \ref{fig:different_rho}, when $\rho\ge40\%$, ElasticTrainer can achieve up to 3$\times$ of training speedup and 2.5$\times$ of FLOPs saving without noticeable accuracy loss, compared to full training. Also, ElasticTrainer can reach up to 2.5$\times$ and 3.5$\times$ more training speedup in wall-clock time compared to BN+Bias and PruneTrain, respectively. On the other hand, when $\rho$ drops to 35\%, the NN accuracy begins to exhibit noticeable reduction. Figure \ref{fig:flops_vs_epoch} further shows that with runtime adaptation, ElasticTrainer can maintain similar FLOPs saving in every epoch to achieve the speedup objective.
	
		\begin{figure}[ht]
		\centering
		\vspace{-0.2in}
		\subfigure[Accuracy on Jetson TX2] { 
			\includegraphics[width=0.22\textwidth]{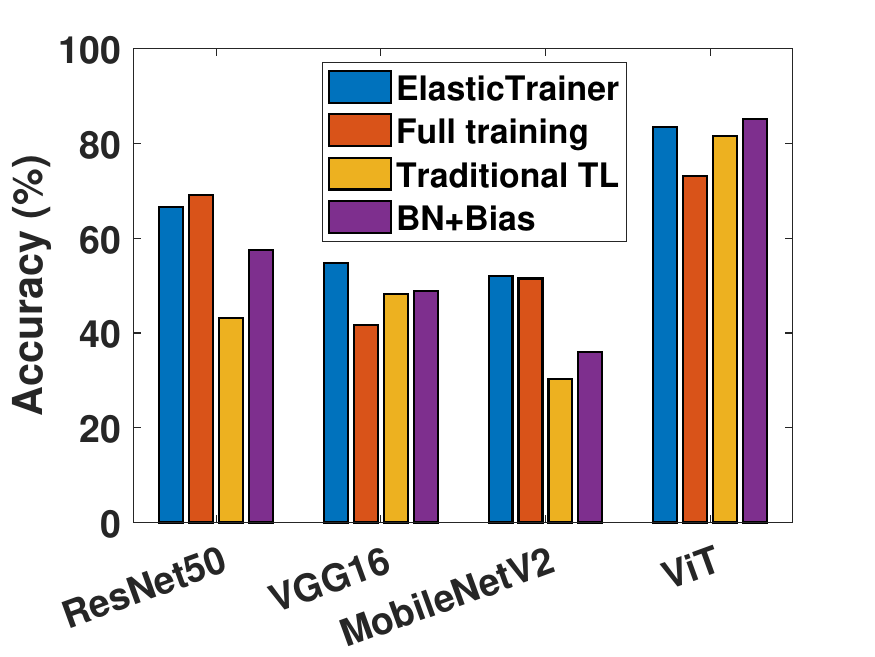}
			\label{fig:different_model_acc_jetson}
		}
		\subfigure[Time on Jetson TX2] { 
			\includegraphics[width=0.22\textwidth]{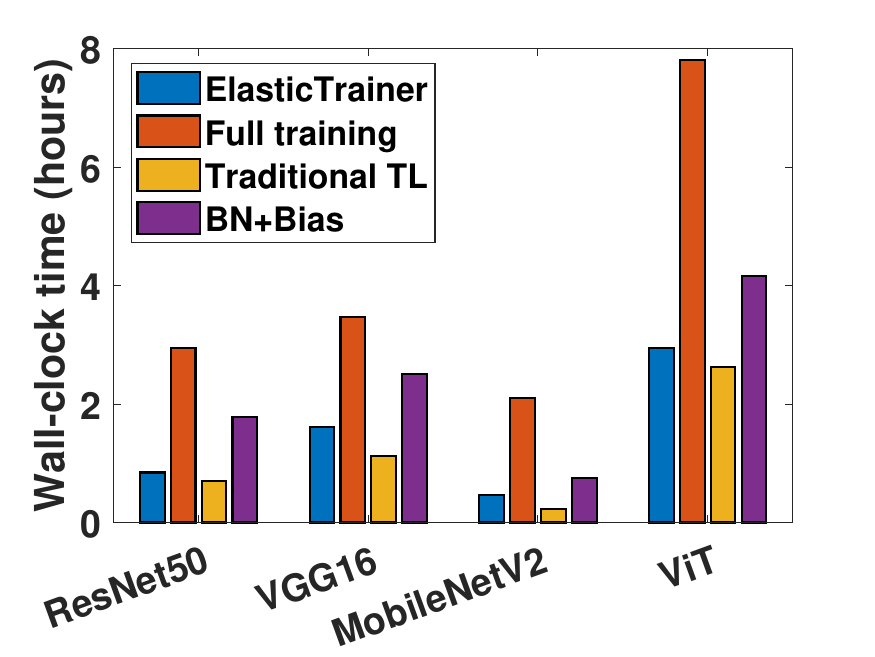}
			\label{fig:different_model_time_jetson}
		}\newline
		
		\vspace{-0.1in}
		\subfigure[Accuracy on Raspberry Pi] { 
			\includegraphics[width=0.22\textwidth]{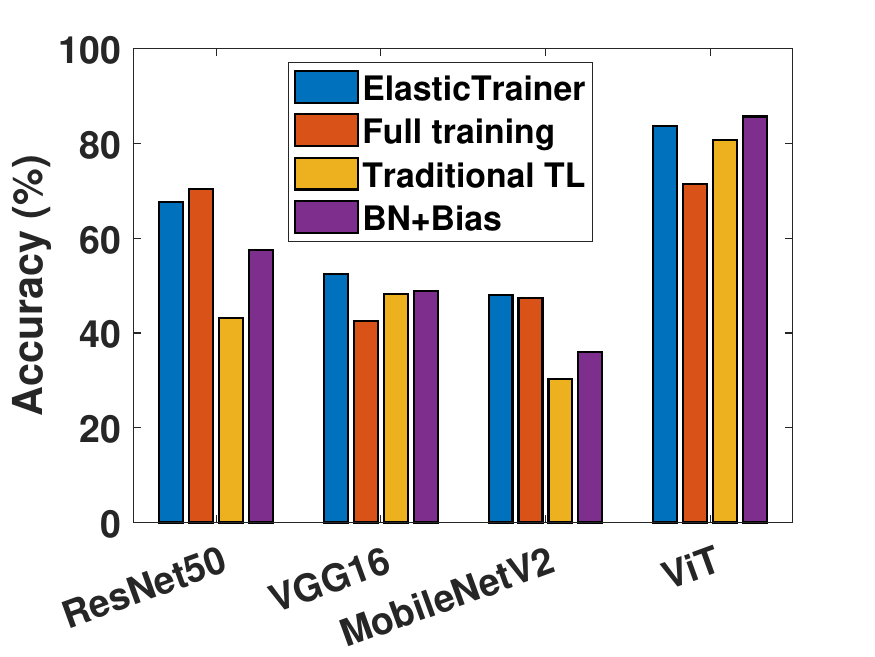}
			\label{fig:different_model_acc_pi}
		}
		\subfigure[Time on Raspberry Pi] { 
			\includegraphics[width=0.22\textwidth]{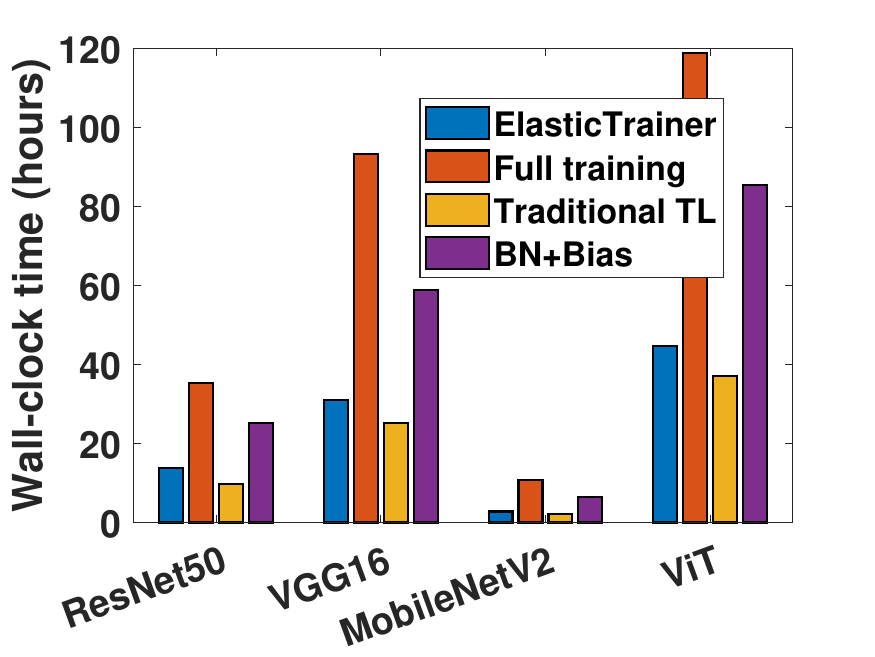}
			\label{fig:different_model_time_pi}
		}
		
		\hspace{-0.4in}
		\vspace{-0.15in}
		\caption{Training different NN models on CUB-200 dataset}
		\label{fig:different_models}
		\vspace{-0.1in}
	\end{figure}

	\begin{figure}
		\centering
		\vspace{-0.1in}
		\hspace{-0.25in}
		\subfigure[CUB-200] { 
			\includegraphics[width=0.24\textwidth]{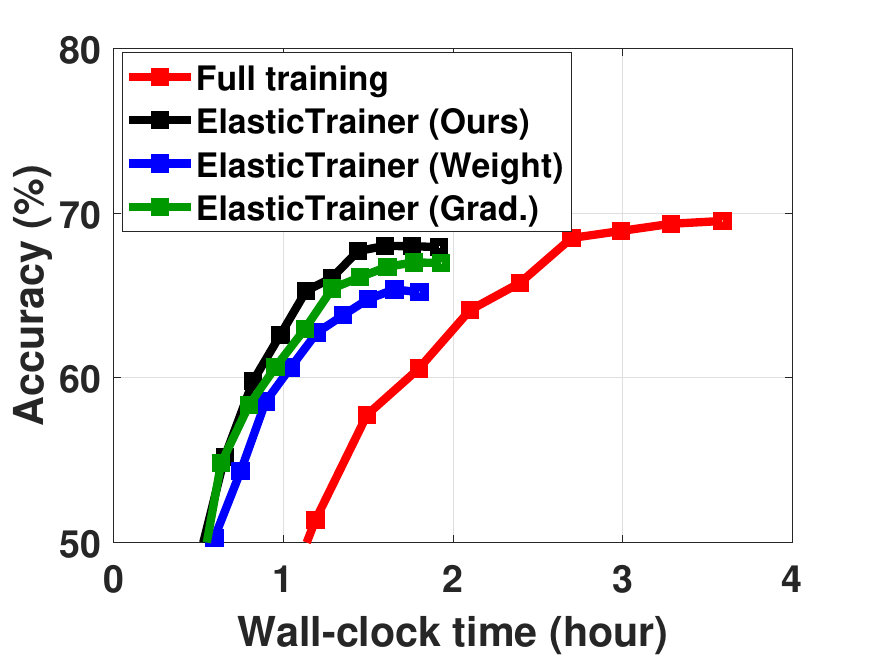}
			\label{fig:importance_evaluator_ablation_cub200_resnet50_jetson}
		}
		\subfigure[Oxford-IIIT Pet] { 
			\includegraphics[width=0.24\textwidth]{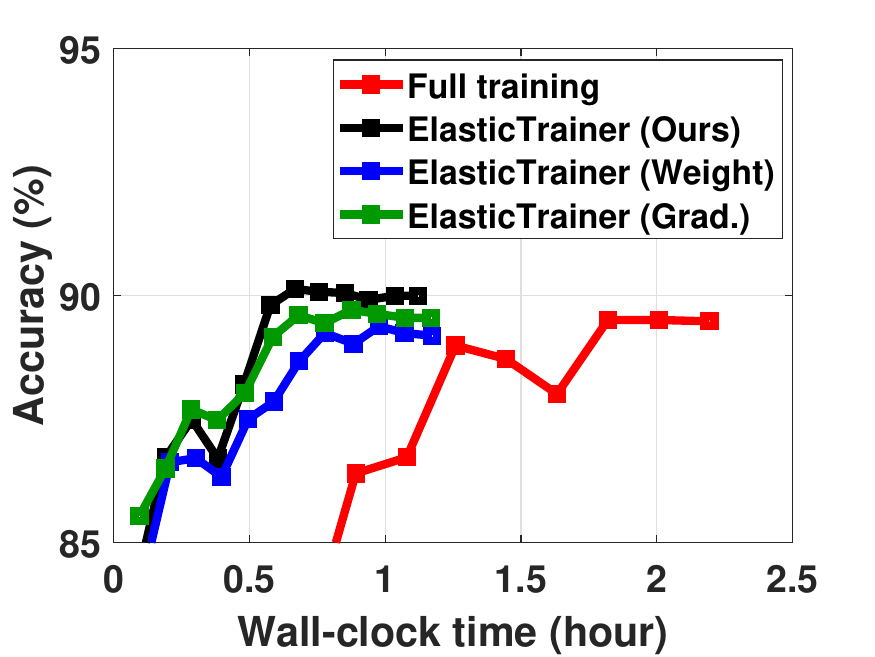}
			\label{fig:importance_evaluator_ablation_oxford_iiit_pet_resnet50_jetson}
		}
		\hspace{-0.4in}
		\vspace{-0.15in}
		\caption{The effectiveness of different tensor importance metrics}
		\label{fig:importance_ablation}
		\vspace{-0.15in}
	\end{figure}
	
		\begin{figure*}[ht]
		\centering
		\hspace{0.05in}
		\subfigure[ResNet50 on CUB-200 dataset, $\rho=70\%$] { 
			\includegraphics[width=0.45\textwidth]{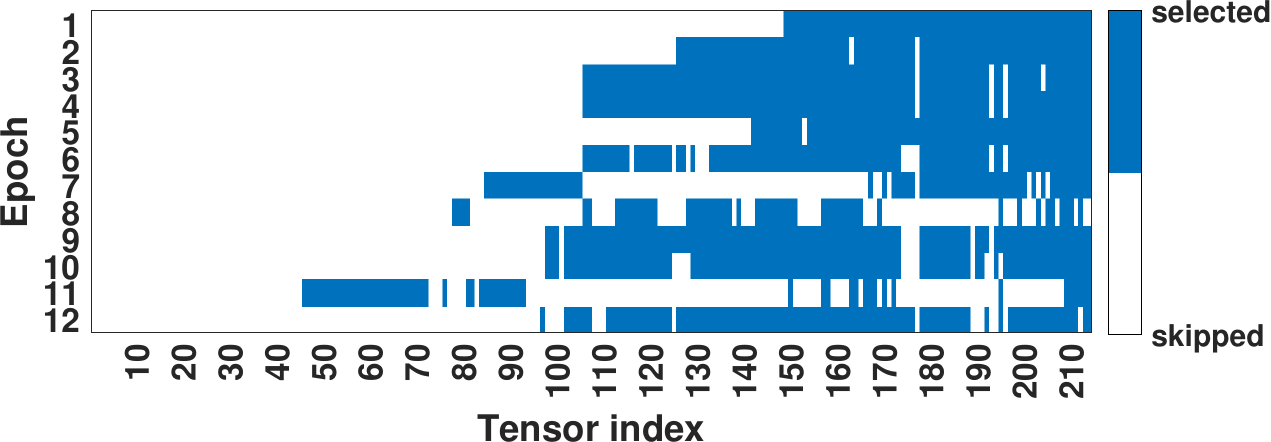}
			\label{fig:mask_logs_resnet50_rho70_cub200}
		}
		\hspace{0.1in}
		\subfigure[ResNet50 on CUB-200 dataset, $\rho=50\%$] { 
			\includegraphics[width=0.45\textwidth]{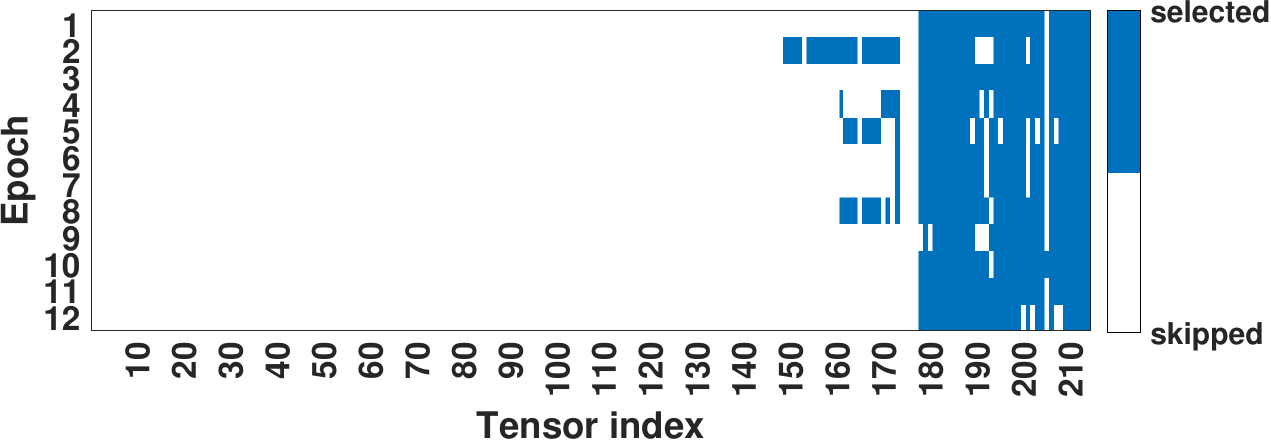}
			\label{fig:mask_logs_resnet50_rho50_cub200}
		}\newline
		
		\vspace{-0.1in}
		\subfigure[ResNet50 on Oxford-IIIT Pet dataset, $\rho=70\%$] { 
			\includegraphics[width=0.45\textwidth]{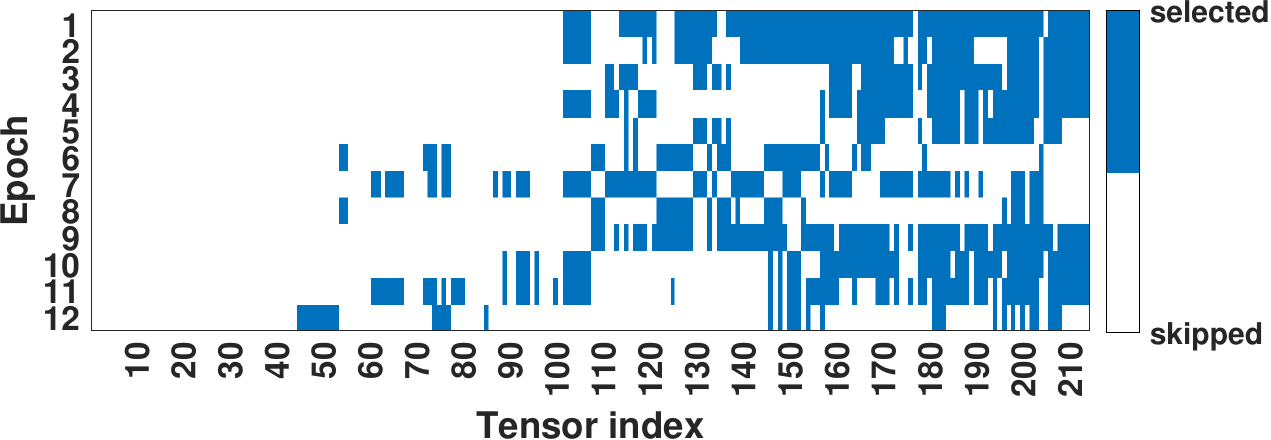}
			\label{fig:mask_logs_resnet50_rho70_pet37}
		}
		\hspace{0.05in}
		\subfigure[VGG16 on Oxford-IIIT Pet dataset, $\rho=70\%$] { 
			\includegraphics[width=0.45\textwidth]{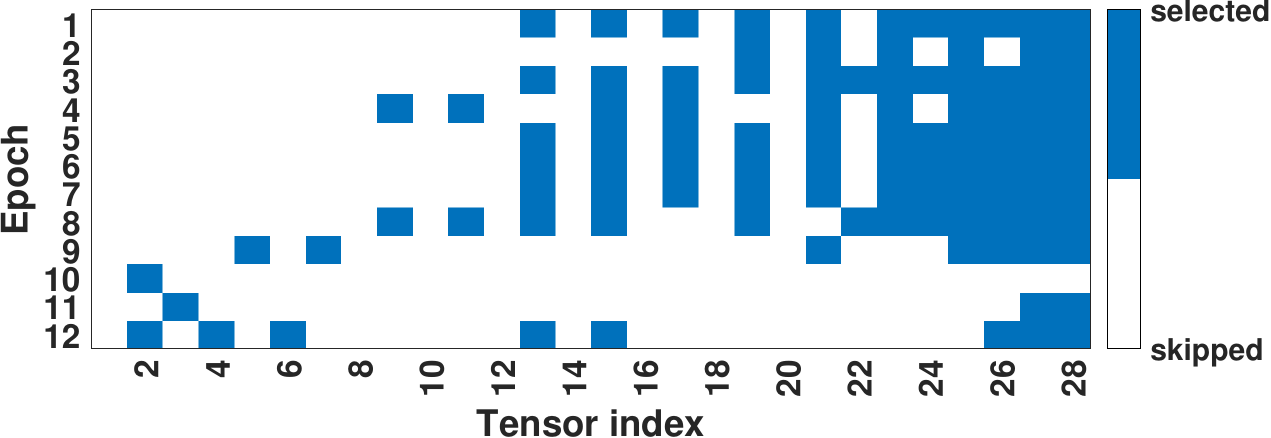}
			\label{fig:mask_logs_vgg16_rho70_pet37}
		}
		
		\hspace{-0.4in}
		\vspace{-0.2in}
		\caption{Elastic tensor selections in different training epochs}
		\label{fig:selection_strategy}
		\vspace{-0.15in}
	\end{figure*}
	
	Note that, being different from wall-clock time, FLOPs ignores the heterogeneity hardware accelerations for different NN operations. Hence, although PruneTrain can save $>$30\% in training FLOPs, it spends a significant amount of time to compute the extra loss functions for weight sparsification and reshaping NN structures at runtime \cite{lym2019prunetrain}, which result in  little wall-clock time saving.


	\vspace{-0.05in}
	\subsection{Performance on Different NN Models}
	The learning efficiency could vary on different NN models. Over-parametrized models (e.g., VGG16 and ViT) can potentially achieve higher accuracy but could easily experience overfitting. Lightweight models (e.g., MobileNetV2) can run backward pass faster but may have lower accuracy. We use ElasticTrainer to train three different NN models with the training speedup objective $\rho$=40\%. Results in Figure \ref{fig:different_models} show that ElasticTrainer achieves 2$\times$-4$\times$ speedup compared to full training and up to 2.1$\times$ more speedup compared to other existing schemes, without noticeable accuracy loss. 
	
	In particular, we found that training a complex model (e.g., ResNet50) with our scheme of elastic tensor selection is even faster than fully training a much smaller model (e.g., MobileNetV2), and smaller models such as MobileNetV2 converge poorly with the 12 training epochs due to its insufficient model complexity. As a result, in our experiments we run a sufficient number of extra epochs until it converges. In these cases, ElasticTrainer could speed up MobileNetV2's training by 4$\times$.
	
	Besides, ElasticTrainer improves the final accuracy of VGG16's training by $>$10\% even compared to full training. This is because VGG16 is a plain convolutional NN without advanced substructures (e.g., residual connection \cite{he2016deep} used in ResNet50 and MobileNetV2) that improves learning efficiency. Instead, ElasticTrainer improves its convergence and prevents overfitting by only selecting the important tensors for training.

	\subsection{Efficacy of Tensor Importance Evaluator}
	The model accuracy that can be achieved by ElasticTrainer directly relates to the tensor importance metric being used. As mentioned in Section 4, the additivity of our proposed metric allows those most important tensors to be fairly selected and ensures that such selection will not be biased by different value scales. To demonstrate its effectiveness, we compare our proposed metric to the existing metrics that evaluate tensor importance based on weight magnitudes (Weight) \cite{han2015learning} and training feedback (Grad.) \cite{belay2022gradient} but are not additive, by applying all these metrics to our proposed scheme of ElasticTrainer. As shown in Figure \ref{fig:importance_ablation}, our tensor importance evaluator helps ElasticTrainer maintain high accuracy over the entire training process, and achieve 1\%-5\% higher final accuracy than the existing metrics, with the ResNet50 model on CUB-200 and Oxford-IIIT Pet datasets.

	\subsection{Behavior of Elastic Tensor Selection}
	The performance of ElasticTrainer is ensured by elastic tensor selection at runtime, and we analyzed such selection behaviors with different NN models, datasets and speedup objectives. In general, as shown in Figure \ref{fig:selection_strategy}, top tensors (lower indices) are less likely to be selected due to the high time cost of passing error gradients. If top tensors are indeed very important and should be selected, ElasticTrainer can adaptively skip updating\footnote{The FLOPs of updating each tensor (i.e., computing weight updates) can be as significant as the FLOPs of passing error gradients through this tensor} a few bottom tensors (higher indices) with lower importance at the same time to retain the desired training speedup.
	
	Also, tensor selection tends to be more consecutive on more difficult datasets (CUB-200) and more sparse on simpler datasets (Oxford-IIIT Pet). This is because two adjacent tensors tend to have coupled functionality in feature extraction, and should always be both retrained if any one is selected. By further comparing Figure \ref{fig:mask_logs_resnet50_rho70_cub200} with \ref{fig:mask_logs_resnet50_rho50_cub200}, higher speedup ratio $\rho$ could help remove such restriction in selection and the elasticity becomes more significant. 

	\begin{figure}[ht]
		\centering
		\vspace{-0.2in}
		\hspace{-0.25in}
		\subfigure[Jetson TX2] { 
			\includegraphics[width=0.24\textwidth]{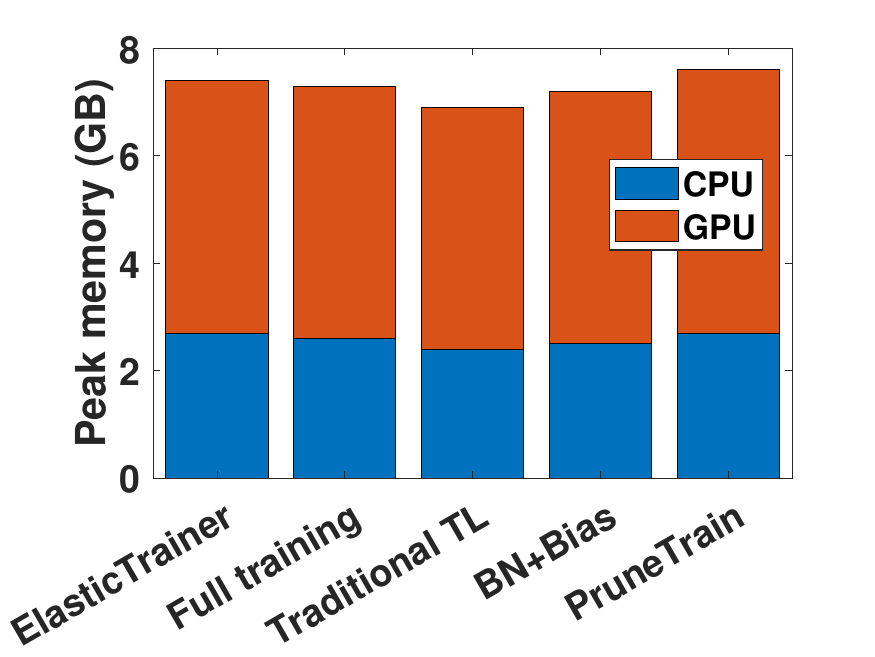}
			\label{fig:memory_jetson}
		}
		\subfigure[Raspberry Pi 4B] { 
			\includegraphics[width=0.24\textwidth]{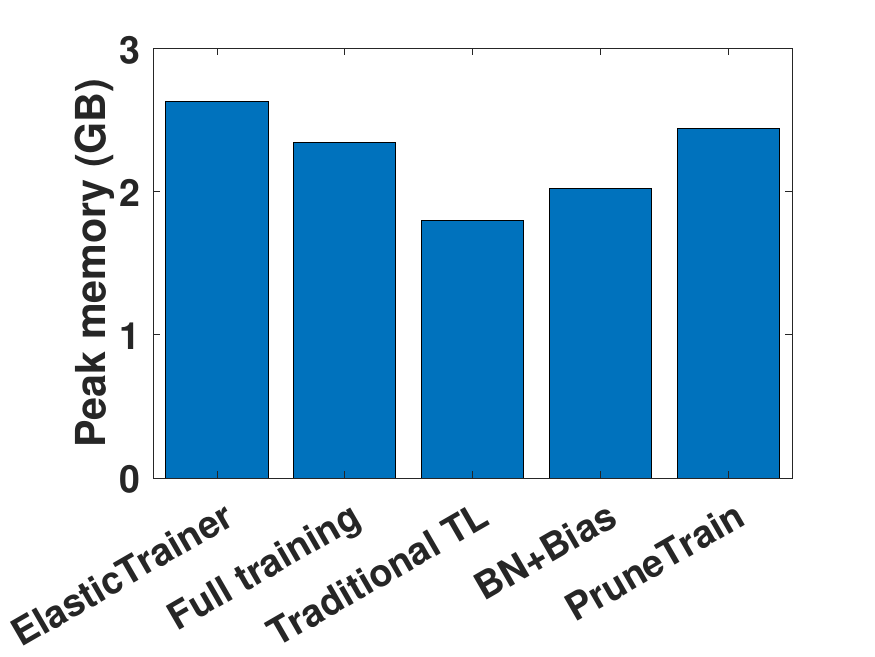}
			\label{fig:memory_pi}
		}
		\hspace{-0.4in}
		\vspace{-0.15in}
		\caption{Memory cost on different devices}
		\label{fig:memory}
		\vspace{-0.15in}
	\end{figure}
	
	\vspace{-0.05in}
	\subsection{Memory Consumption}
	NN training is usually memory intensive, due to the need of buffering intermediate features and training feedback. We use jetson-stats \cite{jtop} and htop \cite{htop} as profiling tools to measure the peak memory usage of ElasticTrainer during training on Jetson TX2 and Raspberry Pi 4B, respectively. Since the CPU and GPU on Jetson TX2 share the same memory, we separately report CPU and GPU memory usage. As shown in Figure \ref{fig:memory_jetson}, on Jetson TX2, ElasticTrainer's memory consumption is comparable to all other schemes, and its CPU memory usage is slightly higher than full training due to the limitation of TensorFlow, which re-generates computing graph after each tensor selection but doesn't free the memory used by the previous computing graph. Such memory leakage becomes more significant on Raspberry Pi devices with only CPUs, where ElasticTrainer consumes 10\% more memory than full training. However, such memory consumption still remains well within the memory capacity of embedded devices, and we expect that such memory leakage can be easily fixed via software reprogramming.
	
	\begin{figure}[ht]
		\centering
		\vspace{-0.15in}
		\hspace{-0.25in}
		\subfigure[Jetson TX2] { 
			\includegraphics[width=0.24\textwidth]{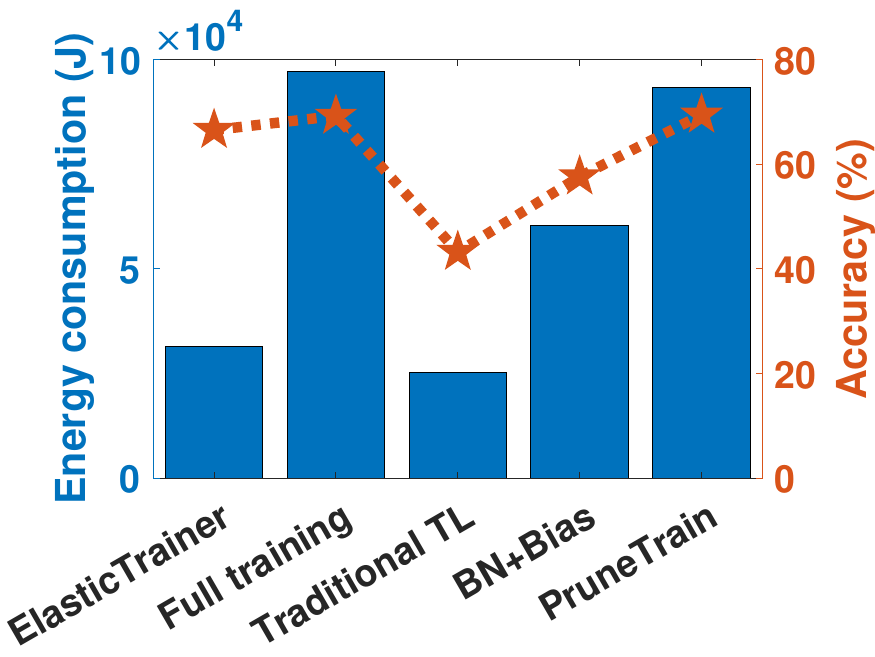}
			\label{fig:energy_jetson}
		}
		\subfigure[Raspberry Pi 4B] { 
			\includegraphics[width=0.24\textwidth]{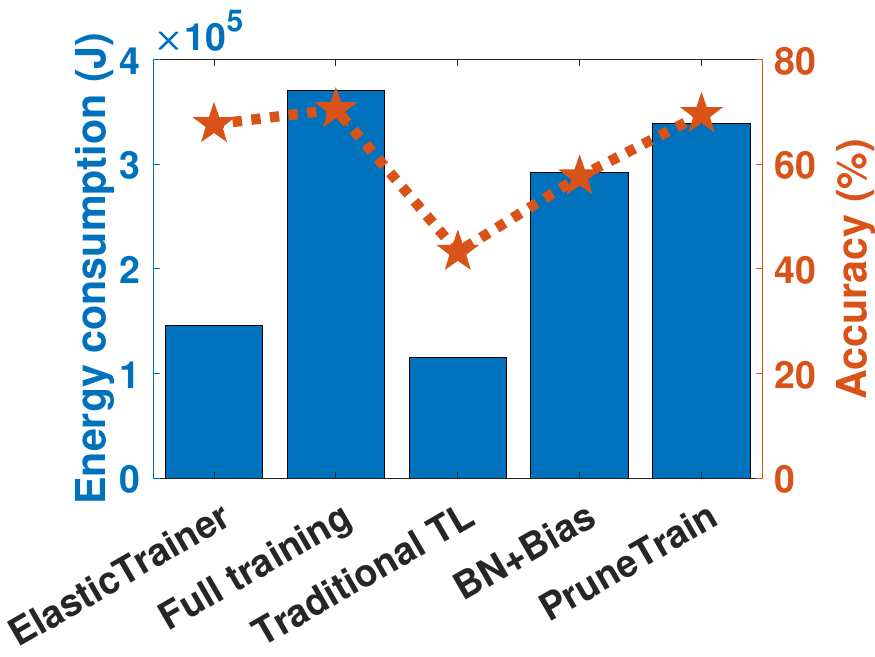}
			\label{fig:energy_pi}
		}
		\hspace{-0.4in}
		\vspace{-0.1in}
		\caption{Energy cost on different devices}
		\label{fig:energy}
	\end{figure}
	
	\vspace{-0.05in}
	\subsection{Energy consumption}
	Continuous on-device training can be energy-consuming, and we use a Ponnie power meter \cite{poniie} to measure its energy consumption. As shown in Figure \ref{fig:energy}, compared to full training, ElasticTrainer reduces the energy consumption by up to 3$\times$ without noticeable accuracy loss. Similarly, all the other schemes have lower energy efficiency, because they have either high energy cost (BN+Bias and PruneTrain) or low NN accuracy (Traditional TL).

	\section{Related Work}
	\noindent\textbf{On-device transfer learning.}
	The design of ElasticTrainer builds on the concept of transfer learning. Traditional transfer learning \cite{donahue2014decaf, sharif2014cnn} either trains only the classification module \cite{cai2020tinytl}, or limits the trainable NN potion to specific types of NN components \cite{zaken2021bitfit,lin2022device,mudrakarta2018k}. However, since the selection of trainable NN portion is always done offline, it has limited adaptability to new online data.
	
	\noindent \textbf{Dynamic neural networks.} ElasticTrainer is related to dynamic NNs \cite{yoon2017lifelong}, including both model pruning and growth techniques that selectively memorize knowledge from the streaming data. However, they mainly aim to overcome catastrophic forgetting \cite{kirkpatrick2017overcoming} instead of speeding up on-device training. Further, dynamic NNs usually require extra training efforts to reduce the inconsistency among NN structures, and hence achieve little training speedup in practical settings. For example, pruning methods require an additional finetuning stage to minimize the accuracy loss \cite{molchanov2019importance}. Model growth schemes need to train the newly added NN layers from scratch and cannot adopt pre-trained models. They hence have higher difficulty in training and only apply to simple learning tasks such as digit recognition \cite{deng2012mnist}. 
	
	\noindent \textbf{Approximate training.}
	ElasticTrainer's tensor selection approach is related to recent techniques of approximate training. Most existing work reduces training computation by either quantizing \cite{alistarh2017qsgd, goli2020resprop} or cutting off \cite{sun2017meprop} the propagation of error gradients at certain neurons, but accumulatively weakens the training feedback passed to earlier layers and leads to low accuracy. Alternatively, other schemes propose to selectively skip less important training data \cite{mourad2019online}, but it only applies to simple datasets with large redundancy. In many practical scenarios, the majority of data samples can be valuable to help the NN learn to capture new patterns, and hence should not be simply discarded.


	\section{Discussions}
	\noindent\textbf{More accurate tensor time model.}
	Some NN training software \cite{xu2018cavs} supports runtime graph optimization for better memory efficiency. Such runtime variations of the NN computing graph can impact the accuracy of our tensor time model being built offline. In that case, ElasticTrainer can reactively rebuild the tensor time model by profiling the NN at runtime, to adapt to the computing graph changes. Further, current profiling tools could introduce significant memory overhead at runtime, and need to balance between profiling frequency and time model's adaptability.
        
        \noindent\textbf{Improving evaluation of tensor importance.}
ElasticTrainer's tensor importance metric is calculated based on the training loss, which may not precisely reflect each tensor's contribution when overfitting exists in the later stage of training. To address this problem, one possible solution is to adopt more informative metrics (e.g., based on validation loss) as the indicator to the model accuracy. Although some existing schemes \cite{gao2021network, joo2022metrics} provided methods to approximately calculate such validation accuracy and loss, they are too computationally expensive for runtime use. Exploring more computationally efficient solutions will be our future work.
        
        
	\noindent\textbf{Generality of ElasticTrainer.}
In this paper, we implement and evaluate ElasticTrainer's performance using TensorFlow \cite{abadi2016tensorflow} on Jetson and Raspberry Pi devices. ElasticTrainer can also be deployed to other computing platforms, such as smartphones. The potential difficulty is that some AI frameworks (e.g., TensorFlow Lite \cite{tflite} and MNN \cite{mnn}) on smartphones don't support dynamic computing graphs at runtime, hence preventing elastic tensor selection. However, many others (e.g., PyTorch \cite{paszke2019pytorch}, MXNet \cite{mxnet}, Chainer \cite{chainer}, and DyNet \cite{dynet}) support dynamic computing graphs, and it is possible to migrate these frameworks to mobile and embedded platforms with engineering efforts.
	
	Similarly, ElasticTrainer can also be applied to other popular NN training diagrams, such as self-supervised learning \cite{devlin2018bert}, reinforcement learning (RL) \cite{arulkumaran2017deep} and federated learning (FL) \cite{bonawitz2019towards}. In these training diagrams, the tensor importance metric may need specific adaptations, such as incorporating the communication cost in FL. Besides, the tensor timing model should also adapt to possible interactions between subnetworks, such as actor-critic structures in RL. Expanding ElasticTrainer to these training diagrams will be our future work.


	\section{Conclusion}
	In this paper, we present ElasticTrainer, a new technique that allows fully elastic selection of NN tensors to adapt to the runtime need of training, so as to always achieve fast and accurate on-device training. ElasticTrainer achieves 3.5$\times$ more speedup in wall-clock time without noticeable accuracy loss when compared to the existing schemes, and also reduces the energy consumption by 2$\times$-3$\times$ more.
	
	\section*{Acknowledgments}
	We thank the anonymous shepherd and reviewers for their comments and feedback. This work was supported in part by National
	Science Foundation (NSF) under grant number CNS-1812407, CNS-2029520, IIS-1956002, IIS-2205360, CCF-2217003 and CCF-2215042.
	
	
	\appendix
	\section*{Artifact Appendix}
	The research artifacts accompanying this paper are available via https://doi.org/10.5281/zenodo.7812233 (for source codes and Raspberry Pi 4 system image) and https://doi.org/10.5281/zenodo.7812218 (for Nvidia Jetson TX2 system image).

	\bibliographystyle{abbrv}
	\bibliography{ref}

\begin{thebibliography}{10}

\bibitem{chainer}
Chainer.
\newblock \url{https://github.com/chainer/chainer}.

\bibitem{dynet}
Dynet.
\newblock \url{https://github.com/clab/dynet}.

\bibitem{htop}
htop.
\newblock \url{https://htop.dev/}.

\bibitem{jetbot}
Jetbot.
\newblock \url{https://developer.nvidia.com/embedded/learn/jetbot}.

\bibitem{jtop}
jetson-stats.
\newblock \url{https://github.com/rbonghi/jetson_stats}.

\bibitem{mnn}
Mnn.
\newblock \url{https://github.com/alibaba/MNN}.

\bibitem{mxnet}
Mxnet.
\newblock \url{https://github.com/apache/mxnet}.

\bibitem{jetsontx2}
Nvidia jetson tx2.
\newblock \url{https://developer.nvidia.com/embedded/jetson-tx2}.

\bibitem{poniie}
Poniie power meter.
\newblock \url{https://poniie.com/products/6}.

\bibitem{rpi4}
Raspberry pi 4b.
\newblock \url{https://www.raspberrypi.com/products/raspberry-pi-4-model-b/}.

\bibitem{skydio}
Skydio 2.
\newblock \url{https://developer.nvidia.com/blog/skydio-2-jetson-tx2-drone/}.

\bibitem{tflite}
Tensorflow lite.
\newblock \url{https://www.tensorflow.org/lite}.

\bibitem{abadi2016tensorflow}
M.~Abadi.
\newblock Tensorflow: learning functions at scale.
\newblock In {\em Proceedings of the 21st ACM SIGPLAN International Conference
  on Functional Programming}, pages 1--1, 2016.

\bibitem{akata2015evaluation}
Z.~Akata, S.~Reed, D.~Walter, H.~Lee, and B.~Schiele.
\newblock Evaluation of output embeddings for fine-grained image
  classification.
\newblock In {\em Proceedings of the IEEE conference on computer vision and
  pattern recognition}, pages 2927--2936, 2015.

\bibitem{alistarh2017qsgd}
D.~Alistarh, D.~Grubic, J.~Li, R.~Tomioka, and M.~Vojnovic.
\newblock Qsgd: Communication-efficient sgd via gradient quantization and
  encoding.
\newblock {\em Advances in neural information processing systems}, 30, 2017.

\bibitem{amari1993backpropagation}
S.-i. Amari.
\newblock Backpropagation and stochastic gradient descent method.
\newblock {\em Neurocomputing}, 5(4-5):185--196, 1993.

\bibitem{amodei2016deep}
D.~Amodei, S.~Ananthanarayanan, R.~Anubhai, J.~Bai, E.~Battenberg, C.~Case,
  J.~Casper, B.~Catanzaro, Q.~Cheng, G.~Chen, et~al.
\newblock Deep speech 2: End-to-end speech recognition in english and mandarin.
\newblock In {\em International conference on machine learning}, pages
  173--182. PMLR, 2016.

\bibitem{arulkumaran2017deep}
K.~Arulkumaran, M.~P. Deisenroth, M.~Brundage, and A.~A. Bharath.
\newblock Deep reinforcement learning: A brief survey.
\newblock {\em IEEE Signal Processing Magazine}, 34(6):26--38, 2017.

\bibitem{ba2016layer}
J.~L. Ba, J.~R. Kiros, and G.~E. Hinton.
\newblock Layer normalization.
\newblock {\em arXiv preprint arXiv:1607.06450}, 2016.

\bibitem{bahdanau2014neural}
D.~Bahdanau, K.~Cho, and Y.~Bengio.
\newblock Neural machine translation by jointly learning to align and
  translate.
\newblock {\em arXiv preprint arXiv:1409.0473}, 2014.

\bibitem{baldi2013understanding}
P.~Baldi and P.~J. Sadowski.
\newblock Understanding dropout.
\newblock {\em Advances in neural information processing systems}, 26, 2013.

\bibitem{baydin2018automatic}
A.~G. Baydin, B.~A. Pearlmutter, A.~A. Radul, and J.~M. Siskind.
\newblock Automatic differentiation in machine learning: a survey.
\newblock {\em Journal of Marchine Learning Research}, 18:1--43, 2018.

\bibitem{belay2022gradient}
K.~Belay.
\newblock Gradient and mangitude based pruning for sparse deep neural networks.
\newblock In {\em Proceedings of the AAAI Conference on Artificial
  Intelligence}, volume~36, pages 13126--13127, 2022.

\bibitem{bengio2012deep}
Y.~Bengio.
\newblock Deep learning of representations for unsupervised and transfer
  learning.
\newblock In {\em Proceedings of ICML workshop on unsupervised and transfer
  learning}, pages 17--36. JMLR Workshop and Conference Proceedings, 2012.

\bibitem{bonawitz2019towards}
K.~Bonawitz, H.~Eichner, W.~Grieskamp, D.~Huba, A.~Ingerman, V.~Ivanov,
  C.~Kiddon, J.~Kone{\v{c}}n{\`y}, S.~Mazzocchi, B.~McMahan, et~al.
\newblock Towards federated learning at scale: System design.
\newblock {\em Proceedings of Machine Learning and Systems}, 1:374--388, 2019.

\bibitem{breiman2001random}
L.~Breiman.
\newblock Random forests.
\newblock {\em Machine learning}, 45(1):5--32, 2001.

\bibitem{brown2020language}
T.~Brown, B.~Mann, N.~Ryder, M.~Subbiah, J.~D. Kaplan, P.~Dhariwal,
  A.~Neelakantan, P.~Shyam, G.~Sastry, A.~Askell, et~al.
\newblock Language models are few-shot learners.
\newblock {\em Advances in neural information processing systems},
  33:1877--1901, 2020.

\bibitem{cai2020tinytl}
H.~Cai, C.~Gan, L.~Zhu, and S.~Han.
\newblock Tinytl: Reduce activations, not trainable parameters for efficient
  on-device learning.
\newblock {\em arXiv preprint arXiv:2007.11622}, 2020.

\bibitem{chatfield2014return}
K.~Chatfield, K.~Simonyan, A.~Vedaldi, and A.~Zisserman.
\newblock Return of the devil in the details: Delving deep into convolutional
  nets.
\newblock {\em arXiv preprint arXiv:1405.3531}, 2014.

\bibitem{chen2015net2net}
T.~Chen, I.~Goodfellow, and J.~Shlens.
\newblock Net2net: Accelerating learning via knowledge transfer.
\newblock {\em arXiv preprint arXiv:1511.05641}, 2015.

\bibitem{deng2009imagenet}
J.~Deng, W.~Dong, R.~Socher, L.-J. Li, K.~Li, and L.~Fei-Fei.
\newblock Imagenet: A large-scale hierarchical image database.
\newblock In {\em 2009 IEEE conference on computer vision and pattern
  recognition}, pages 248--255. Ieee, 2009.

\bibitem{deng2012mnist}
L.~Deng.
\newblock The mnist database of handwritten digit images for machine learning
  research [best of the web].
\newblock {\em IEEE signal processing magazine}, 29(6):141--142, 2012.

\bibitem{devlin2018bert}
J.~Devlin, M.-W. Chang, K.~Lee, and K.~Toutanova.
\newblock Bert: Pre-training of deep bidirectional transformers for language
  understanding.
\newblock {\em arXiv preprint arXiv:1810.04805}, 2018.

\bibitem{donahue2014decaf}
J.~Donahue, Y.~Jia, O.~Vinyals, J.~Hoffman, N.~Zhang, E.~Tzeng, and T.~Darrell.
\newblock Decaf: A deep convolutional activation feature for generic visual
  recognition.
\newblock In {\em International conference on machine learning}, pages
  647--655. PMLR, 2014.

\bibitem{dosovitskiy2020image}
A.~Dosovitskiy, L.~Beyer, A.~Kolesnikov, D.~Weissenborn, X.~Zhai,
  T.~Unterthiner, M.~Dehghani, M.~Minderer, G.~Heigold, S.~Gelly, et~al.
\newblock An image is worth 16x16 words: Transformers for image recognition at
  scale.
\newblock {\em arXiv preprint arXiv:2010.11929}, 2020.

\bibitem{everingham2010pascal}
M.~Everingham, L.~Van~Gool, C.~K. Williams, J.~Winn, and A.~Zisserman.
\newblock The pascal visual object classes (voc) challenge.
\newblock {\em International journal of computer vision}, 88(2):303--338, 2010.

\bibitem{gao2021network}
S.~Gao, F.~Huang, W.~Cai, and H.~Huang.
\newblock Network pruning via performance maximization.
\newblock In {\em Proceedings of the IEEE/CVF Conference on Computer Vision and
  Pattern Recognition}, pages 9270--9280, 2021.

\bibitem{gim2022memory}
I.~Gim and J.~Ko.
\newblock Memory-efficient dnn training on mobile devices.
\newblock In {\em Proceedings of the 20th Annual International Conference on
  Mobile Systems, Applications and Services}, pages 464--476, 2022.

\bibitem{goli2020resprop}
N.~Goli and T.~M. Aamodt.
\newblock Resprop: Reuse sparsified backpropagation.
\newblock In {\em Proceedings of the IEEE/CVF Conference on Computer Vision and
  Pattern Recognition}, pages 1548--1558, 2020.

\bibitem{guo2020adafilter}
Y.~Guo, Y.~Li, L.~Wang, and T.~Rosing.
\newblock Adafilter: Adaptive filter fine-tuning for deep transfer learning.
\newblock In {\em Proceedings of the AAAI Conference on Artificial
  Intelligence}, volume~34, pages 4060--4066, 2020.

\bibitem{han2015learning}
S.~Han, J.~Pool, J.~Tran, and W.~Dally.
\newblock Learning both weights and connections for efficient neural network.
\newblock {\em Advances in neural information processing systems}, 28, 2015.

\bibitem{he2016deep}
K.~He, X.~Zhang, S.~Ren, and J.~Sun.
\newblock Deep residual learning for image recognition.
\newblock In {\em Proceedings of the IEEE conference on computer vision and
  pattern recognition}, pages 770--778, 2016.

\bibitem{he2017channel}
Y.~He, X.~Zhang, and J.~Sun.
\newblock Channel pruning for accelerating very deep neural networks.
\newblock In {\em Proceedings of the IEEE international conference on computer
  vision}, pages 1389--1397, 2017.

\bibitem{hecht1992theory}
R.~Hecht-Nielsen.
\newblock Theory of the backpropagation neural network.
\newblock In {\em Neural networks for perception}, pages 65--93. Elsevier,
  1992.

\bibitem{hu2021lora}
E.~J. Hu, Y.~Shen, P.~Wallis, Z.~Allen-Zhu, Y.~Li, S.~Wang, L.~Wang, and
  W.~Chen.
\newblock Lora: Low-rank adaptation of large language models.
\newblock {\em arXiv preprint arXiv:2106.09685}, 2021.

\bibitem{hu2015face}
G.~Hu, Y.~Yang, D.~Yi, J.~Kittler, W.~Christmas, S.~Z. Li, and T.~Hospedales.
\newblock When face recognition meets with deep learning: an evaluation of
  convolutional neural networks for face recognition.
\newblock In {\em Proceedings of the IEEE international conference on computer
  vision workshops}, pages 142--150, 2015.

\bibitem{ioffe2015batch}
S.~Ioffe and C.~Szegedy.
\newblock Batch normalization: Accelerating deep network training by reducing
  internal covariate shift.
\newblock In {\em International conference on machine learning}, pages
  448--456. PMLR, 2015.

\bibitem{irsoy2019continuously}
O.~Irsoy and E.~Alpayd{\i}n.
\newblock Continuously constructive deep neural networks.
\newblock {\em IEEE transactions on neural networks and learning systems},
  31(4):1124--1133, 2019.

\bibitem{jeong2022band}
J.~S. Jeong, J.~Lee, D.~Kim, C.~Jeon, C.~Jeong, Y.~Lee, and B.-G. Chun.
\newblock Band: coordinated multi-dnn inference on heterogeneous mobile
  processors.
\newblock In {\em Proceedings of the 20th Annual International Conference on
  Mobile Systems, Applications and Services}, pages 235--247, 2022.

\bibitem{jiang2022model}
Y.~Jiang, S.~Wang, V.~Valls, B.~J. Ko, W.-H. Lee, K.~K. Leung, and
  L.~Tassiulas.
\newblock Model pruning enables efficient federated learning on edge devices.
\newblock {\em IEEE Transactions on Neural Networks and Learning Systems},
  2022.

\bibitem{joo2022metrics}
D.~Joo, S.~Baek, and J.~Kim.
\newblock Which metrics for network pruning: Final accuracy? or accuracy drop?
\newblock In {\em 2022 IEEE International Conference on Image Processing
  (ICIP)}, pages 1071--1075. IEEE, 2022.

\bibitem{kemker2018measuring}
R.~Kemker, M.~McClure, A.~Abitino, T.~Hayes, and C.~Kanan.
\newblock Measuring catastrophic forgetting in neural networks.
\newblock In {\em Proceedings of the AAAI Conference on Artificial
  Intelligence}, volume~32, 2018.

\bibitem{khosla2011novel}
A.~Khosla, N.~Jayadevaprakash, B.~Yao, and F.-F. Li.
\newblock Novel dataset for fine-grained image categorization: Stanford dogs.
\newblock In {\em Proc. CVPR workshop on fine-grained visual categorization
  (FGVC)}, volume~2. Citeseer, 2011.

\bibitem{kingma2014adam}
D.~P. Kingma and J.~Ba.
\newblock Adam: A method for stochastic optimization.
\newblock {\em arXiv preprint arXiv:1412.6980}, 2014.

\bibitem{kirkpatrick2017overcoming}
J.~Kirkpatrick, R.~Pascanu, N.~Rabinowitz, J.~Veness, G.~Desjardins, A.~A.
  Rusu, K.~Milan, J.~Quan, T.~Ramalho, A.~Grabska-Barwinska, et~al.
\newblock Overcoming catastrophic forgetting in neural networks.
\newblock {\em Proceedings of the national academy of sciences},
  114(13):3521--3526, 2017.

\bibitem{krizhevsky2009learning}
A.~Krizhevsky, G.~Hinton, et~al.
\newblock Learning multiple layers of features from tiny images.
\newblock 2009.

\bibitem{kumar2021rma}
A.~Kumar, Z.~Fu, D.~Pathak, and J.~Malik.
\newblock Rma: Rapid motor adaptation for legged robots.
\newblock {\em arXiv preprint arXiv:2107.04034}, 2021.

\bibitem{li2021hermes}
A.~Li, J.~Sun, P.~Li, Y.~Pu, H.~Li, and Y.~Chen.
\newblock Hermes: an efficient federated learning framework for heterogeneous
  mobile clients.
\newblock In {\em Proceedings of the 27th Annual International Conference on
  Mobile Computing and Networking}, pages 420--437, 2021.

\bibitem{li2020look}
C.~Li, S.~Ge, D.~Zhang, and J.~Li.
\newblock Look through masks: Towards masked face recognition with de-occlusion
  distillation.
\newblock In {\em Proceedings of the 28th ACM International Conference on
  Multimedia}, pages 3016--3024, 2020.

\bibitem{li2022pyramidfl}
C.~Li, X.~Zeng, M.~Zhang, and Z.~Cao.
\newblock Pyramidfl: A fine-grained client selection framework for efficient
  federated learning.
\newblock In {\em Proceedings of the 28th Annual International Conference on
  Mobile Computing And Networking}, pages 158--171, 2022.

\bibitem{li2016pruning}
H.~Li, A.~Kadav, I.~Durdanovic, H.~Samet, and H.~P. Graf.
\newblock Pruning filters for efficient convnets.
\newblock {\em arXiv preprint arXiv:1608.08710}, 2016.

\bibitem{lin2022device}
J.~Lin, L.~Zhu, W.-M. Chen, W.-C. Wang, C.~Gan, and S.~Han.
\newblock On-device training under 256kb memory.
\newblock {\em arXiv preprint arXiv:2206.15472}, 2022.

\bibitem{loshchilov2017decoupled}
I.~Loshchilov and F.~Hutter.
\newblock Decoupled weight decay regularization.
\newblock {\em arXiv preprint arXiv:1711.05101}, 2017.

\bibitem{lym2019prunetrain}
S.~Lym, E.~Choukse, S.~Zangeneh, W.~Wen, S.~Sanghavi, and M.~Erez.
\newblock Prunetrain: fast neural network training by dynamic sparse model
  reconfiguration.
\newblock In {\em Proceedings of the International Conference for High
  Performance Computing, Networking, Storage and Analysis}, pages 1--13, 2019.

\bibitem{molchanov2019importance}
P.~Molchanov, A.~Mallya, S.~Tyree, I.~Frosio, and J.~Kautz.
\newblock Importance estimation for neural network pruning.
\newblock In {\em Proceedings of the IEEE/CVF Conference on Computer Vision and
  Pattern Recognition}, pages 11264--11272, 2019.

\bibitem{mourad2019online}
S.~Mourad, H.~Vikalo, and A.~Tewfik.
\newblock Online selective training for faster neural network learning.
\newblock In {\em 2019 IEEE Data Science Workshop (DSW)}, pages 135--139. IEEE,
  2019.

\bibitem{mudrakarta2018k}
P.~K. Mudrakarta, M.~Sandler, A.~Zhmoginov, and A.~Howard.
\newblock K for the price of 1: Parameter-efficient multi-task and transfer
  learning.
\newblock {\em arXiv preprint arXiv:1810.10703}, 2018.

\bibitem{parkhi2012cats}
O.~M. Parkhi, A.~Vedaldi, A.~Zisserman, and C.~Jawahar.
\newblock Cats and dogs.
\newblock In {\em 2012 IEEE conference on computer vision and pattern
  recognition}, pages 3498--3505. IEEE, 2012.

\bibitem{paszke2019pytorch}
A.~Paszke, S.~Gross, F.~Massa, A.~Lerer, J.~Bradbury, G.~Chanan, T.~Killeen,
  Z.~Lin, N.~Gimelshein, L.~Antiga, et~al.
\newblock Pytorch: An imperative style, high-performance deep learning library.
\newblock {\em Advances in neural information processing systems}, 32, 2019.

\bibitem{poirot2019split}
M.~G. Poirot, P.~Vepakomma, K.~Chang, J.~Kalpathy-Cramer, R.~Gupta, and
  R.~Raskar.
\newblock Split learning for collaborative deep learning in healthcare.
\newblock {\em arXiv preprint arXiv:1912.12115}, 2019.

\bibitem{sandler2018mobilenetv2}
M.~Sandler, A.~Howard, M.~Zhu, A.~Zhmoginov, and L.-C. Chen.
\newblock Mobilenetv2: Inverted residuals and linear bottlenecks.
\newblock In {\em Proceedings of the IEEE conference on computer vision and
  pattern recognition}, pages 4510--4520, 2018.

\bibitem{selvaraju2017grad}
R.~R. Selvaraju, M.~Cogswell, A.~Das, R.~Vedantam, D.~Parikh, and D.~Batra.
\newblock Grad-cam: Visual explanations from deep networks via gradient-based
  localization.
\newblock In {\em Proceedings of the IEEE international conference on computer
  vision}, pages 618--626, 2017.

\bibitem{sharif2014cnn}
A.~Sharif~Razavian, H.~Azizpour, J.~Sullivan, and S.~Carlsson.
\newblock Cnn features off-the-shelf: an astounding baseline for recognition.
\newblock In {\em Proceedings of the IEEE conference on computer vision and
  pattern recognition workshops}, pages 806--813, 2014.

\bibitem{simonyan2014very}
K.~Simonyan and A.~Zisserman.
\newblock Very deep convolutional networks for large-scale image recognition.
\newblock {\em arXiv preprint arXiv:1409.1556}, 2014.

\bibitem{smith2022legged}
L.~Smith, J.~C. Kew, X.~B. Peng, S.~Ha, J.~Tan, and S.~Levine.
\newblock Legged robots that keep on learning: Fine-tuning locomotion policies
  in the real world.
\newblock In {\em 2022 International Conference on Robotics and Automation
  (ICRA)}, pages 1593--1599. IEEE, 2022.

\bibitem{sun2017meprop}
X.~Sun, X.~Ren, S.~Ma, and H.~Wang.
\newblock meprop: Sparsified back propagation for accelerated deep learning
  with reduced overfitting.
\newblock In {\em International Conference on Machine Learning}, pages
  3299--3308. PMLR, 2017.

\bibitem{sundararajan2017axiomatic}
M.~Sundararajan, A.~Taly, and Q.~Yan.
\newblock Axiomatic attribution for deep networks.
\newblock In {\em International conference on machine learning}, pages
  3319--3328. PMLR, 2017.

\bibitem{tan2008improving}
G.~Tan, N.~Sun, and G.~R. Gao.
\newblock Improving performance of dynamic programming via parallelism and
  locality on multicore architectures.
\newblock {\em IEEE Transactions on Parallel and Distributed Systems},
  20(2):261--274, 2008.

\bibitem{torralba2011unbiased}
A.~Torralba and A.~A. Efros.
\newblock Unbiased look at dataset bias.
\newblock In {\em CVPR 2011}, pages 1521--1528. IEEE, 2011.

\bibitem{turki2022mega}
H.~Turki, D.~Ramanan, and M.~Satyanarayanan.
\newblock Mega-nerf: Scalable construction of large-scale nerfs for virtual
  fly-throughs.
\newblock In {\em Proceedings of the IEEE/CVF Conference on Computer Vision and
  Pattern Recognition}, pages 12922--12931, 2022.

\bibitem{vaswani2017attention}
A.~Vaswani, N.~Shazeer, N.~Parmar, J.~Uszkoreit, L.~Jones, A.~N. Gomez,
  {\L}.~Kaiser, and I.~Polosukhin.
\newblock Attention is all you need.
\newblock {\em Advances in neural information processing systems}, 30, 2017.

\bibitem{vepakomma2018split}
P.~Vepakomma, O.~Gupta, T.~Swedish, and R.~Raskar.
\newblock Split learning for health: Distributed deep learning without sharing
  raw patient data.
\newblock {\em arXiv preprint arXiv:1812.00564}, 2018.

\bibitem{WahCUB_200_2011}
C.~Wah, S.~Branson, P.~Welinder, P.~Perona, and S.~Belongie.
\newblock The caltech-ucsd birds-200-2011 dataset.
\newblock Technical Report CNS-TR-2011-001, California Institute of Technology,
  2011.

\bibitem{wang2011approximation}
X.~Wang, W.~Wu, and D.~Zhu.
\newblock An approximation algorithm for nonlinear 0-1 integer programming
  problems.
\newblock In {\em 2011 International Conference on Computer and Management
  (CAMAN)}, pages 1--5. IEEE, 2011.

\bibitem{wu2018group}
Y.~Wu and K.~He.
\newblock Group normalization.
\newblock In {\em Proceedings of the European conference on computer vision
  (ECCV)}, pages 3--19, 2018.

\bibitem{xu2022mandheling}
D.~Xu, M.~Xu, Q.~Wang, S.~Wang, Y.~Ma, K.~Huang, G.~Huang, X.~Jin, and X.~Liu.
\newblock Mandheling: mixed-precision on-device dnn training with dsp
  offloading.
\newblock In {\em Proceedings of the 28th Annual International Conference on
  Mobile Computing And Networking}, pages 214--227, 2022.

\bibitem{xu2018deeptype}
M.~Xu, F.~Qian, Q.~Mei, K.~Huang, and X.~Liu.
\newblock Deeptype: On-device deep learning for input personalization service
  with minimal privacy concern.
\newblock {\em Proceedings of the ACM on Interactive, Mobile, Wearable and
  Ubiquitous Technologies}, 2(4):1--26, 2018.

\bibitem{xu2018cavs}
S.~Xu, H.~Zhang, G.~Neubig, W.~Dai, J.~K. Kim, Z.~Deng, Q.~Ho, G.~Yang, and
  E.~P. Xing.
\newblock Cavs: An efficient runtime system for dynamic neural networks.
\newblock In {\em 2018 USENIX Annual Technical Conference (USENIX ATC 18)},
  pages 937--950, 2018.

\bibitem{yoon2017lifelong}
J.~Yoon, E.~Yang, J.~Lee, and S.~J. Hwang.
\newblock Lifelong learning with dynamically expandable networks.
\newblock {\em arXiv preprint arXiv:1708.01547}, 2017.

\bibitem{zaken2021bitfit}
E.~B. Zaken, S.~Ravfogel, and Y.~Goldberg.
\newblock Bitfit: Simple parameter-efficient fine-tuning for transformer-based
  masked language-models.
\newblock {\em arXiv preprint arXiv:2106.10199}, 2021.

\bibitem{zhang2020mdldroidlite}
Y.~Zhang, T.~Gu, and X.~Zhang.
\newblock Mdldroidlite: A release-and-inhibit control approach to
  resource-efficient deep neural networks on mobile devices.
\newblock In {\em Proceedings of the 18th Conference on Embedded Networked
  Sensor Systems}, pages 463--475, 2020.

\bibitem{zhu2015aligning}
Y.~Zhu, R.~Kiros, R.~Zemel, R.~Salakhutdinov, R.~Urtasun, A.~Torralba, and
  S.~Fidler.
\newblock Aligning books and movies: Towards story-like visual explanations by
  watching movies and reading books.
\newblock In {\em Proceedings of the IEEE international conference on computer
  vision}, pages 19--27, 2015.

\end{thebibliography}

\end{document}